\documentclass[journal]{IEEEtran}
\usepackage[pagebackref=true,breaklinks=true,letterpaper=true,colorlinks,bookmarks=false,linkcolor=black,citecolor=black,urlcolor=blue]{hyperref}

\usepackage{ifpdf}
\usepackage{cite}
\ifCLASSINFOpdf
   \usepackage[pdftex]{graphicx}
\else
\fi
\usepackage[cmex10]{amsmath}
\usepackage{algorithm}
\usepackage{algorithmic}
\usepackage{array}
\ifCLASSOPTIONcompsoc
  \usepackage[caption=false,font=normalsize,labelfont=sf,textfont=sf]{subfig}
\else
  \usepackage[caption=false,font=footnotesize]{subfig}
\fi
\usepackage{stfloats}
\usepackage{url}
\usepackage{multirow}
\usepackage{graphicx}
\usepackage{algorithmic}
\usepackage{afterpage}
\usepackage{pbox}
\usepackage{enumitem}
\usepackage{color}
\usepackage{amsfonts}

\newcolumntype{C}[1]{>{\centering\let\newline\\\arraybackslash\hspace{0pt}}m{#1}}

\begin{document}
\title{Con-Patch: When a Patch Meets its Context}

\author{Yaniv~Romano,
        and~Michael~Elad,~\IEEEmembership{Fellow,~IEEE}
\thanks{Y. Romano is with the Department of Electrical Engineering, Technion -- Israel Institute of Technology, Technion City, Haifa 32000, Israel. E-mail address: yromano@tx.technion.ac.il. M. Elad  is with the Department of Computer Science, Technion -- Israel Institute of Technology, Technion City, Haifa 32000, Israel. E-mail address: elad@cs.technion.ac.il.}}

\def\x{{\mathbf x}}
\def\Y{{\mathbf Y}}
\def\V{{\mathbf V}}
\def\v{{\mathbf v}}
\def\q{{\mathbf q}}
\def\xh{{\hat{\x}}}
\def\qh{{\hat{\q}}}
\def\xi{\x_i}

\def\oneb{\mathrm{1}}
\def\one{\underbar{$ \oneb $}}
\def\y{{\mathbf y}}
\def\z{{\mathbf z}}
\def\p{{\mathbf p}}
\def\q{{\mathbf q}}
\def\U{{\mathbf U}}

\def\Gam{{\mathbf \Gamma}}
\def\Q{{\mathbf Q}}
\def\D{{\mathbf D}}
\def\E{{\mathbf E}}
\def\W{{\mathbf W}}
\def\Z{{\mathbf Z}}
\def\Zh{{\mathbf \hat \Z }}
\def\V{{\mathbf V}}
\def\S{{\mathbf S}}
\def\G{{\mathbf G}}
\def\Dh{{\hat{\D}}}
\def\H{\mathcal{H}}
 
\def\L{{\mathrm L}}
\def\I{{\mathrm I}}
\def\b{{\mathrm b}}
\def\W{{\mathbf W}}
\def\A{{\mathbf A}}
\def\B{{\mathrm B}}
\def\Ai{\A_i}
\def\T{{\mathrm T}}
\def\Ti{\T_i}
\def\R{{\mathbf R}}
\def\P{{\mathbf P}}
\def\r{{\mathrm r}}
\def\c{{\mathrm c}}
\def\hhsigma{{\hat{\sigma}}}
\def\RR{{\mathbb R}}

\maketitle

\begin{abstract}
	Measuring the similarity between patches in images is a fundamental building block in various tasks. Naturally, the patch-size has a major impact on the matching quality, and on the consequent application performance. Under the assumption that our patch database is sufficiently sampled, using large patches (e.g. 21-by-21) should be preferred over small ones (e.g. \mbox{7-by-7}). However, this "dense-sampling" assumption is rarely true; in most cases large patches cannot find relevant nearby examples. This phenomenon is a consequence of the curse of dimensionality, stating that the database-size should grow exponentially with the patch-size to ensure proper matches. This explains the favored choice of small patch-size in most applications.
	
	Is there a way to keep the simplicity and work with small patches while getting some of the benefits that large patches provide? In this work we offer such an approach. We propose to concatenate the regular content of a conventional (small) patch with a compact representation of its (large) surroundings -- its context. Therefore, with a minor increase of the dimensions (e.g. with additional 10 values to the patch representation), we implicitly/softly describe the information of a large patch. The additional descriptors are computed based on a self-similarity behavior of the patch surrounding.
	
	We show that this approach achieves better matches, compared to the use of conventional-size patches, without the need to increase the database-size. Also, the effectiveness of the proposed method is tested on three distinct problems: (i) External natural image denoising, (ii) Depth image super-resolution, and (iii) Motion-compensated frame-rate up-conversion.
\end{abstract}


\IEEEpeerreviewmaketitle

\section{Introduction}
\label{intro}

\IEEEPARstart{P}{atch} matching is extensively used in many image processing applications, including image restoration \cite{levin2011natural, levin2012patch ,zontak2011internal, mosseri2013combining,adaptive2015database,buades2005non,glasner2009super,mac2012patch,kheradmand2014general,romano2014single}, video compression based on motion estimation \cite{furht2012motion,sullivan2012overview,ydar2014fruc}, saliency detection \cite{goferman2012context}, and more.
Consider, for example, the problem of image denoising which will accompany our discussion as a case study; given a noise-corrupted version $ \Y \in \RR^{N} $ of a clean image $ \Z \in \RR^{N} $, the task is to approximate the unknown $\Z$. 
"External Denoising" \cite{levin2011natural, levin2012patch, zontak2011internal, mosseri2013combining,adaptive2015database} is a popular and effective approach for estimating the underlying clean image, where patch matching is the core engine in it. More specifically, external denoising methods (i) break the input image into (possibly overlapping) patches, (ii) clean each noisy patch by applying a weighted average over its $ k  $ nearest neighbors ($ k $-NN), chosen from a large database of clean image patches, and (iii) construct the final image by aggregating the cleaned patches. Intuitively, the more meaningful and accurate the neighbors that are found, the better the restoration performance. Note that external denoising can be considered as an extension of the popular Non Local Means (NLM) algorithm \cite{buades2005non}, often called "Internal Denoising" \cite{zontak2011internal} as it seeks for NN in the image itself rather than in an external database, leaning on the self-similarity property of images.

Given a patch database that is sufficiently sampled (e.g. composed of $ 10^{10} $ clean and small patches), Levin and Nadler \cite{levin2011natural, levin2012patch} show that external denoising of the form described above approximates very well the optimal Bayesian minimum mean squared error (MMSE) estimator of a noisy patch. They also study the effect of the patch-size on the denoising performance, showing how the restoration error improves with the growth of the patch dimension. Interestingly, even with the large database of patches used in this work ($ 10^{10} $ examples), the authors encounter difficulties analyzing the performance of the denoising bound for patches that go beyond $ 7\times 7 $ pixels.

One can conclude from the above that working with large patches rapidly becomes impractical since the size of the database should grow exponentially with the patch-size in order to guarantee appropriate and meaningful nearby examples. This explains why relatively small patches are widely preferred in many applications that lean on patches and their matching. 

Broadly speaking, no matter how good the local treatment is when operating on an image, it is limited in modeling its global nature \cite{zoran2011learning,romano2015boosting}. Therefore, in this context, using large patches is a step towards better image modeling \cite{levin2011natural, levin2012patch, jere2016trainlets}, but may seem as a dead-end avenue, due to the difficulties mentioned above.

In this paper we suggest a novel way to keep the advantage of working with small patches, while gaining from the potential that large patches bring.
Our main contribution is the idea of \emph{context-patch} (short name -- "\emph{con-patch}").
We suggest concatenating every small patch (conventional raw data patch, e.g. of size $ 7 \times 7 $) with a compact representation of its large surroundings, called \emph{context-feature} (e.g. $ 10 $ values that describe a $ 21 \times 21 $ patch).
In other words, the con-patch has the explicit content of a small patch and the implicit encapsulated information of its large surrounding.
The proposed approach bypasses the curse of dimensionality since the context-feature is very compact, leading to a minor increase of the dimensions of a small patch. Thus, it does not require to increase the size of the database we work with.

\begin{figure}[!t]
\centering
\includegraphics[width=3.4in]{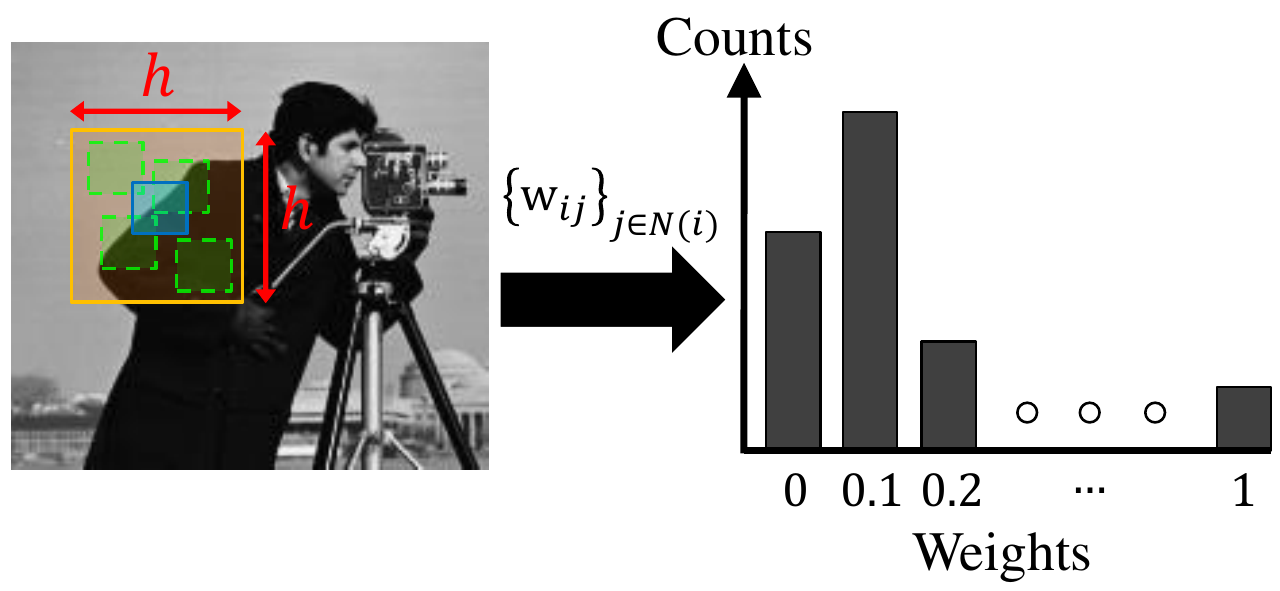}
\caption{The proposed context-feature. First, we measure the similarity weights $ w_{ij} $ between the central patch $ \x_i $ (blue solid line) and its surrounding patches $ \x_j $ (green dashed line) in a window of size $ h \times h $ (orange solid line). Then, the weights are partitioned into a histogram of $ \b $ bins, followed by a normalization step.}
\label{weights_scheme}
\end{figure}

Inspired by the self-similarity descriptor \cite{shechtman2007matching}, we propose a specific context-feature chosen for the augmentation of the patch, based on the self-similarity property between image patches.  This feature-vector expresses how similar a small central patch is to its large surrounding window.
As demonstrated in Fig. \ref{weights_scheme}, we suggest measuring the similarity weights 
\begin{align}
\label{wij}
w_{ij} = \exp{\left\{-\frac{\|\x_i-\x_j\|_2^2}{2\sigma^2}\right\}}, \text{    } \forall j\in \mathcal{N}_h(i)
\end{align}
between a central patch $ \x_i $ and its surrounding patches $ \x_j $ in a neighborhood $ \mathcal{N}_h(i) $ of size $ h \times h $ pixels (called "correlation surface" in \cite{shechtman2007matching}). {In order to reduce computations, instead of having $ h^2 $ different weights, we evaluate $ w_{ij} $ in steps of $ m $ pixels in the horizontal and vertical directions, leading to $ \left(\frac{h}{m}\right)^2 $ weights.}
Then, as a way to reduce the dimensions, the weights are partitioned into a histogram, composed of $ b $ bins. Finally, the histogram is normalized to have a unit sum. Notice that the number of bins serves as the length of the context-feature, independent of the size of the surrounding window, describing some of the information that a large patch/window provide. In order to control the influence of the context-feature on the matching, the normalized histogram is multiplied by a scalar of choice.

It is important to emphasize that working with con-patches is not equivalent to the direct use of large patches. As such, what are the benefits of working with these extended vectors? What is the source of superiority of using the context over the conventional large or small patches? In this paper we show that the context helps by leading to better matches in the NN search; it filters irrelevant and misleading pseudo-matches that bias the search and reduce the application performance. On one hand, thanks to the compact representation of the context, we keep the advantage of working with small patches, for which there are  sufficient meaningful nearby examples in the database. On the other hand, the context guides and ''sharpens'' the search mechanism to matches that are more appropriate by bringing a force that characterizes only large patches.

The concatenation of the context-features to the small patches enables us to easily plug the context in many algorithms, and detach it from the content whenever required.
Furthermore, sticking to the original method's distance measure, as we suggest, does not raise new limitations or additional algorithmic modifications.
For instance, many methods use the $ l_2 $-norm (Euclidean distance) as the similarity measure because of its simplicity \cite{flann_pami_2014}. In this case, plugging the context to the patch content is easily done.

In order to show the effectiveness of the proposed con-patch idea, we focus on several algorithms that rely on a patch matching step. 
We simply replace the conventional patches with our novel con-patches, and show the gained improvement in performance obtained due to the improved matching.
We demonstrate the power of con-patches on three very different applications that expose the generality of the proposed scheme: (i) External denoising, (ii) Single depth image super-resolution (SR), and (iii) Motion compensated frame rate up conversion (MC-FRUC).

Starting with external denoising \cite{zontak2011internal,mosseri2013combining}, as already mentioned, the noise in a given image is reduced by averaging clean matches that are found in the database by a NN search.
The denoising performance is directly influenced by the quality of the neighbors found. With the use of the con-patch, better
matches are obtained, due to the filtering effect of pseudo-matches that mislead the denoising process.
Interestingly, Zontak et al. \cite{zontak2011internal} demonstrate that the performance of internal denoising is better than an  external one\footnote{In a follow-up paper \cite{mosseri2013combining}, it has been shown that a combination of the internal and external approaches outperforms the individual performance of each of the methods.}. This puts our work in its proper perspective, as we show that external denoising with con-patches stir things up and performs much better than the conventional internal or external denoising.

Next, we tackle the SR \cite{mac2012patch} problem, where the task is to increase the resolution of a given depth image.
The authors of \cite{mac2012patch} suggest utilizing a database composed of pairs of low-resolution (LR) and high-resolution (HR) patches.
They search the database for possible HR versions of the input LR patches, then construct the output image by merging the found matches.
This problem enables us to demonstrate the effectiveness of the proposed context-feature for depth images.
In addition, it indicates that the context is beneficial for the prediction of new spatial content.   

The third problem we address with the con-patches is MC-FRUC \cite{ydar2014fruc}, where the task is to increase the frame-rate of a given video. 
The temporal interpolation is achieved by estimating the local motion-trajectories, followed by a motion compensation step.
Patch matching is an inherent part in these methods, where the context plays a key role.
We show that the con-patches have also the ability to improve the motion estimation and thus the prediction of new temporal content.

This paper is organized as follows: In Section \ref{related} we describe several related works in which the idea of context in different settings was proposed and demonstrated. In Section \ref{proposed} we introduce our novel con-patches. Experiments are brought in Section \ref{experiments}, showing an improvement of the proposed idea for external denoising, SR and MC-FRUC. Conclusions are drawn in Section \ref{conclusions}.

\section{Related Work}
\label{related}

There is a large ambiguity around the notion of \emph{context} in the literature; it comes up in many applications and in different ways.
For example, in object recognition there is a major importance to the contextual associations between objects \cite{oliva2007role}, where the context is used in a high-level manner, e.g., by modeling the relations and interactions between objects.
In contrast, in this paper we aim to benefit from the large patch statistics in a compact low-level fashion, putting emphasis on low-level tasks -- image and video restoration.

In the fields of single image SR (i.e. upscaling) \cite{sun2010context,sun2012super} and inpainting (i.e. fill-in missing pixels/holes in an image) \cite{ruzic2015context}, the context is utilized to constrain the search for candidate patches to regions of similar context.
The core idea in these works is to apply a two-stage procedure: (i) seek for large regions with similar contextual behavior, and then (ii) construct the image by integrating patches from the found matched regions.

More specifically, Sun et al. \cite{sun2010context} suggest dividing a database of images into pairs of LR and HR segments (relatively large regions with roughly uniform texture, obtained by applying a segmentation algorithm).
Then, per each input pixel $ p $, the database is searched to find several LR segments that are texturally similar to the input LR image segment (that the pixel $ p $ lies in).
The similarity measure between segments is based on their responses to a derivative filter bank.
Finally, a patch matching step is applied, where the potential nearby patches are constrained to the ones that belong to the matched segments.

In the case of image inpainting, Ruzic et al. \cite{ruzic2015context} suggest dividing the image into variable size windows (segments) and use the responses to Gabor filter bank as a measure of contextual similarity. Then, a Markov random field model is utilized to fill-in the missing pixels by combining patches from the constrained matched windows rather than the whole image.

There are several major differences between \cite{sun2010context, sun2012super, ruzic2015context} and our approach. 
We suggest concatenating the context to the content and apply a ''one-shot'' patch-matching, while the previous works choose a two-step strategy.
In addition, the representation of the context is different -- responses to filter bank applied on segments vs. internal self-similarities of patches in a large window. 
Furthermore, dividing the image into segments with similar textures may be irrelevant for some applications.
For example, in MC-FRUC, when assuming relatively small movements, the motion is estimated by matching patches between two consecutive frames in a limited search window (e.g. $ 21\times 21 $). In this scenario, the NN search is already constrained (the search window can be considered as a matched segment), thus previous works reduce to a conventional patch-matching. In addition, finding segments with similar texture might mislead the matching step, especially when the boundaries of the segment coincide with image structures, or when the textures/edges are deformed due to local motions.

Another related work is the self-similarity descriptor \cite{shechtman2007matching}, which is used in various high-level computer vision tasks, e.g., object recognition.
Similarly to the proposed context-feature, the self-similarity descriptor computes the correlation surface, i.e., the similarity weights between a central patch to its neighbors, as defined in Eq. (\ref{wij}).
Yet, differently from our suggestion, a log-polar mapping is applied on the correlation surface. Finally, a quantization step is applied, leading to a normalized histogram representation of the descriptor.
Our added feature bares some resemblance to the self-similarity descriptor and can be considered as a simpler version of it.
However, different from earlier work, we suggest using the self-similarity descriptor as a way to represent the context of the patch for improving the NN search for low-level image processing applications and it is added to the regular small patch.

We should note that, following our definition of context-feature, any compact representation of the large patch can be considered; the self-similarity feature is one out of many possibilities. 

During the search for effective context-feature, we tested other ways to incorporate the context of the small patch, e.g., by representing the large patch using HoG descriptor \cite{dalal2005histograms}, or by using a downscaled version of the surroundings (which is closely related to the patch foveation operator \cite{foi2012foveated}). Motivated with the popularity of the self-similarity property, which is known to be an effective image prior \cite{buades2005non, shechtman2007matching, zontak2011internal, kheradmand2014general}, we decided to concentrate and demonstrate the great potential of the context by harnessing this feature. In order to have a complete and fair comparison between the self-similarity and other descriptors (e.g. HoG) an extensive parameter tuning should be done (per each descriptor). As such, we decided to omit this comparison from the paper. As a future research direction, it is interesting to test and compare various other structure descriptors.


\begin{figure*}[t]
\centering
  \subfloat[Image]{\includegraphics[width=1.7in]{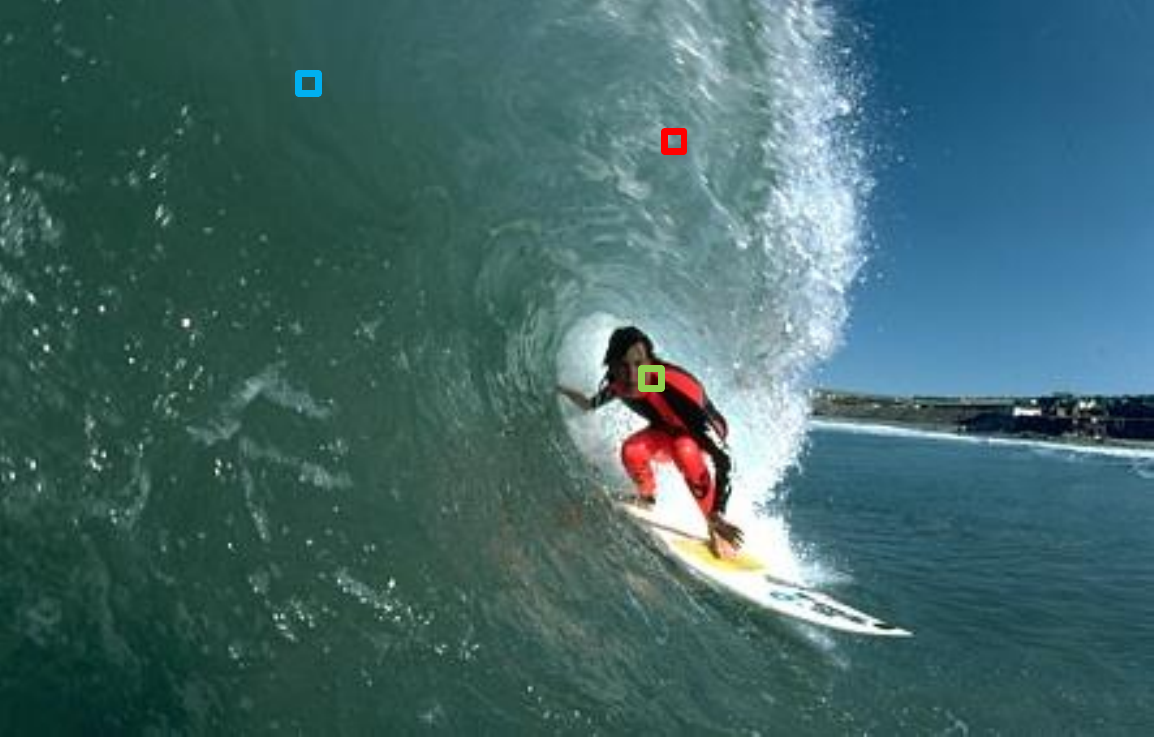}}
  \hfil
  \centering
  \subfloat[$ \H(\x_i) $ of 3 different patches]{\includegraphics[width=1.7in]{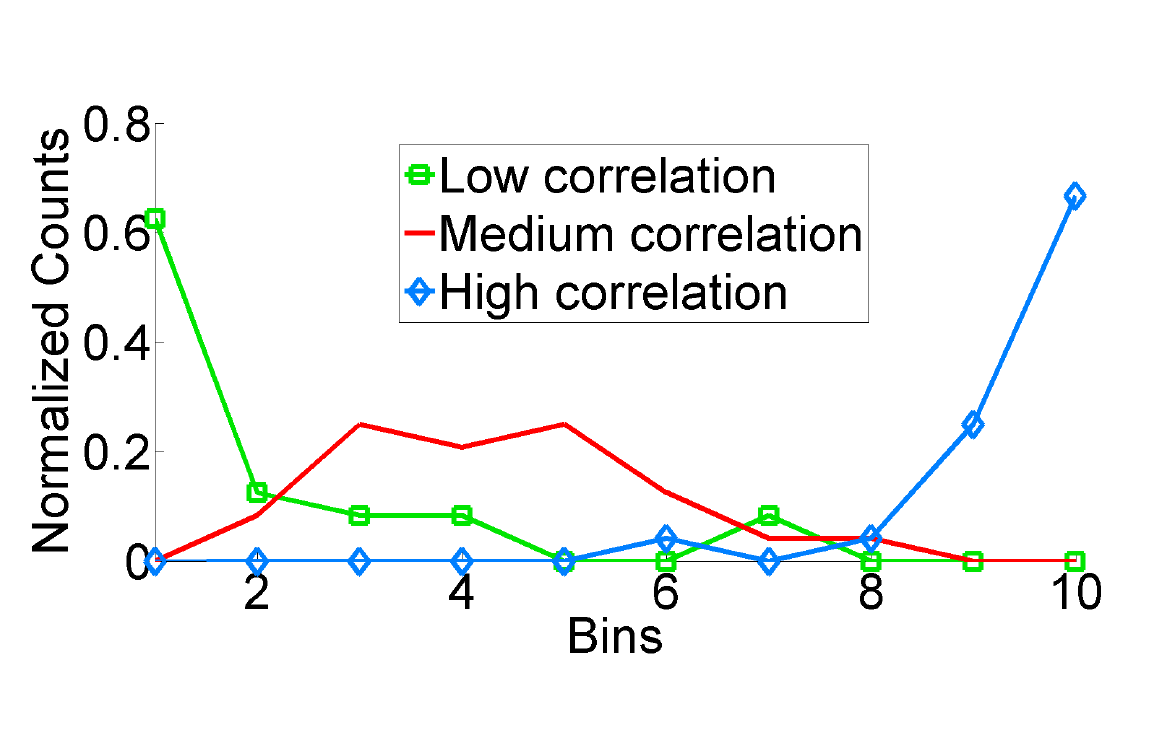}}
  \hfil
  \centering
  \subfloat[Bin \#1 of $ \H(\x_i)$, $\forall i $]{\includegraphics[width=1.7in]{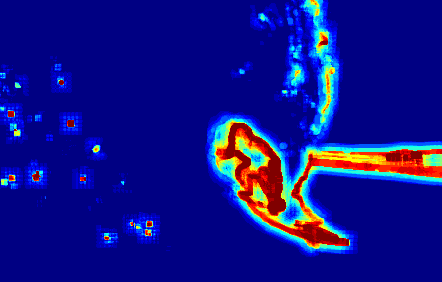}}
  \hfil    
  \centering
  \subfloat[Bin \#2 of $ \H(\x_i)$, $\forall i $]{\includegraphics[width=1.7in]{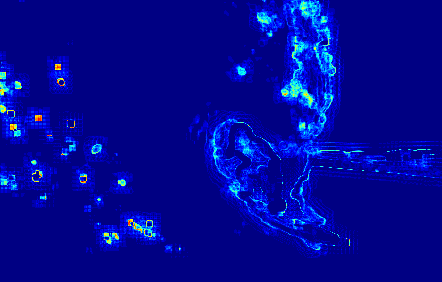}}
  \hfil
  \centering
  \subfloat[Bin \#3 of $ \H(\x_i)$, $\forall i $]{\includegraphics[width=1.7in]{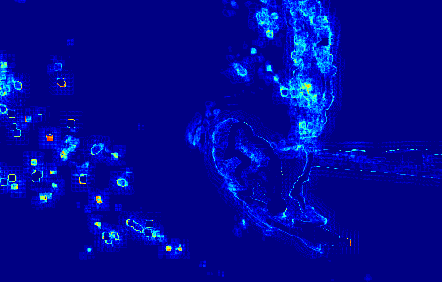}}
  \hfil
  \centering
  \subfloat[Bin \#4 of $ \H(\x_i)$, $\forall i $]{\includegraphics[width=1.7in]{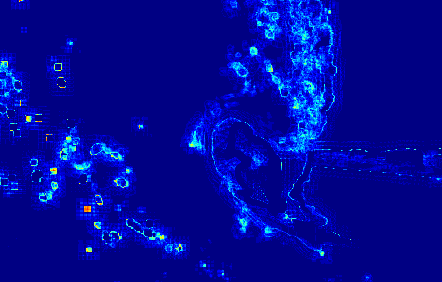}}
    \hfil
    \centering
  \subfloat[Bin \#5 of $ \H(\x_i)$, $\forall i $]{\includegraphics[width=1.7in]{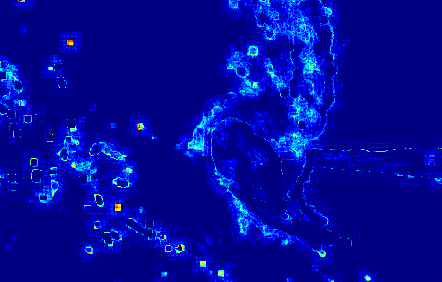}}
    \hfil 
    \centering
  \subfloat[Bin \#6 of $ \H(\x_i)$, $\forall i $]{\includegraphics[width=1.7in]{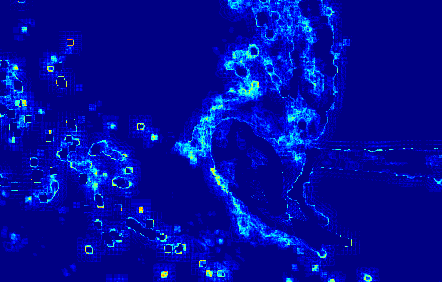}}
    \hfil 
    \centering
  \subfloat[Bin \#7 of $ \H(\x_i)$, $\forall i $]{\includegraphics[width=1.7in]{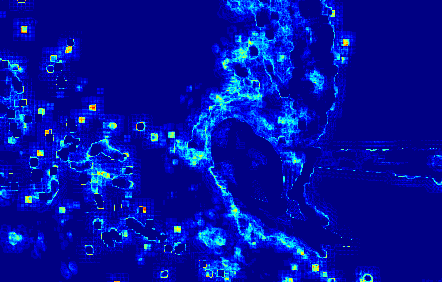}}
    \hfil
    \centering
  \subfloat[Bin \#8 of $ \H(\x_i)$, $\forall i $]{\includegraphics[width=1.7in]{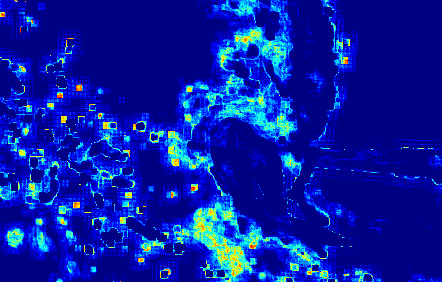}}
    \hfil
    \centering
  \subfloat[Bin \#9 of $ \H(\x_i)$, $\forall i $]{\includegraphics[width=1.7in]{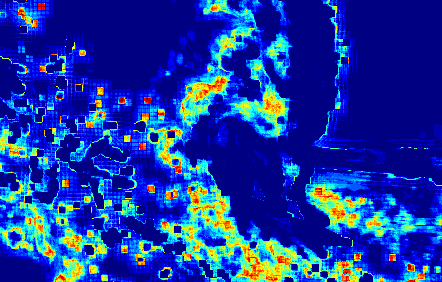}}
    \hfil
    \centering
  \subfloat[Bin \#10 of $ \H(\x_i)$, $\forall i $]{\includegraphics[width=1.7in]{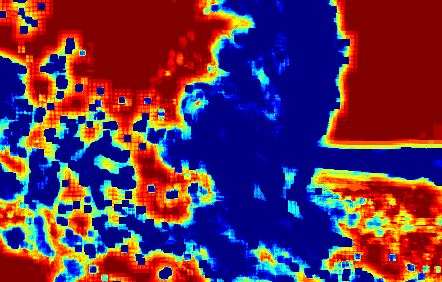}}

\centering
\caption{Visualization of the context-feature, computed on \textsf{Surfer} image. (a) The input image, (b) The histogram $ \H(\x_i) $ of the green, red and blue patches taken from \textsf{Surfer} image, (c)-(l) The values of the normalized histogram $ \H(\x_i) $ per each patch and bin, warm and cold colors indicate low and high values, respectively.}
\label{context_visualization}
\end{figure*}

\begin{figure*}[t]
\centering
  \subfloat[$ \sigma_v = 15 $]{\includegraphics[width=2.2in]{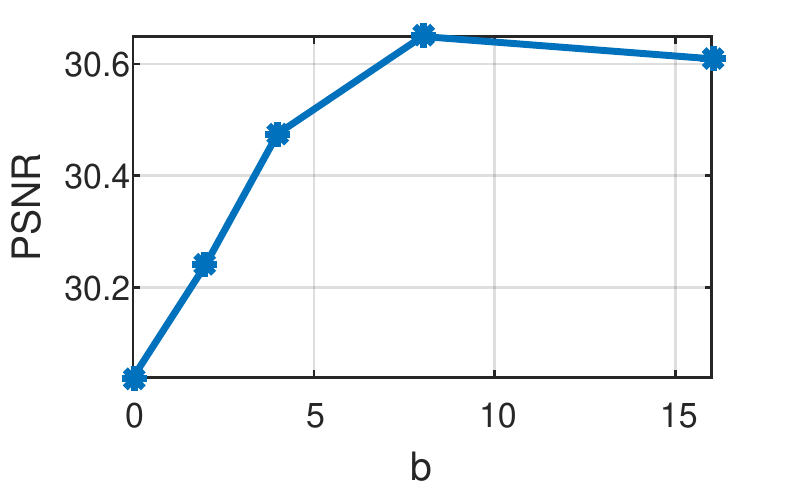}}
  \centering
  \subfloat[$ \sigma_v = 25 $]{\includegraphics[width=2.2in]{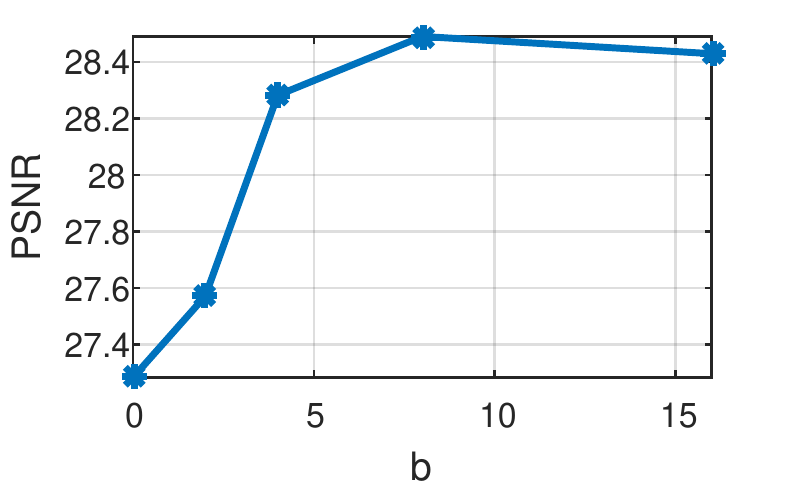}}
  \centering
  \subfloat[$ \sigma_v = 35 $]{\includegraphics[width=2.2in]{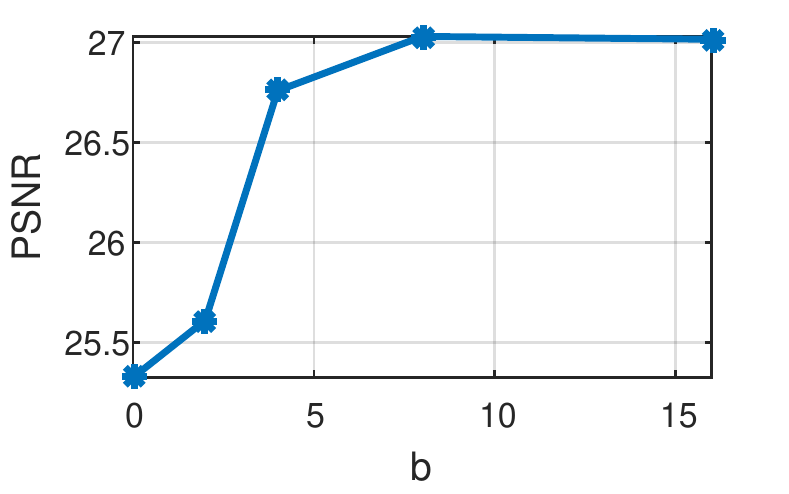}} \\
  \centering
  \subfloat[$ \sigma_v = 50 $]{\includegraphics[width=2.2in]{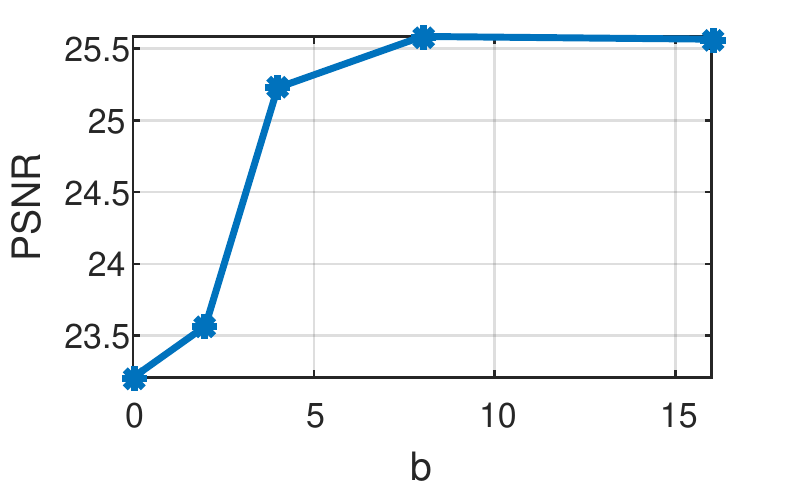}}
  \centering
  \subfloat[$ \sigma_v = 75 $]{\includegraphics[width=2.2in]{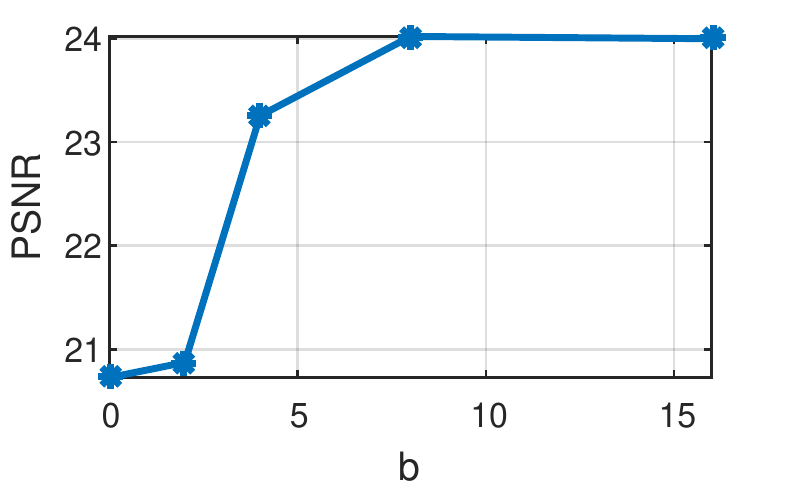}}
  \centering
  \subfloat[$ \sigma_v = 100 $]{\includegraphics[width=2.2in]{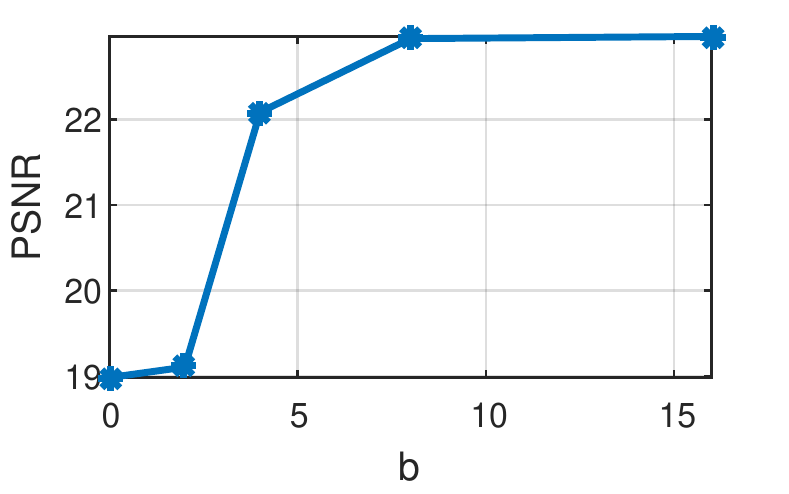}}
    
\centering
\caption{The effect of the length of the context-feature $ \H $ on external denoising performance for different noise-levels, varying from (a) $ \sigma_v = 15 $ up-to (f) $ \sigma_v = 100 $. The results obtained by averaging the PSNR over 10 randomly images, taken from the test-set of BSDS300 database. Per each noise level, the similarity weights are partitioned into $ b $ equally-spaced bins ($ b \in \{0, 2, 4, 8, 16\}$), leading to context-feature of length $ b $. Note that conventional external filtering is obtained when $ b=0 $, serving as a baseline for comparison.}
\label{set_bins}
\end{figure*}

\section{The Proposed Con-Patches}
\label{proposed}
In this section we provide a detailed description of the proposed context-feature. Then, we show how to construct the con-patches.

\subsection{Self-Similarity Feature}
\label{self_similarity}

The core idea behind the context-feature is to represent the large surroundings of a patch in a compact fashion. Once we have a low dimensional feature that describes the large patch, we are able to keep the simplicity and work with small patches while having some of the flavor of working with large ones. Naturally, there are many ways to encapsulate the large patch information. In this paper we rely on the self-similarity property of images that has drawn much attention in recent years. 

Fig. \ref{weights_scheme} illustrates the procedure of computing the context-feature $ \H(\x_i) $, describing some of the information of the large surrounding of the patch $ \x_i $. Following the self-similarity descriptor \cite{shechtman2007matching}, we compute the correlation surface --  set of similarity weights between a central patch $ \x_i $ and its surrounding patches $ \x_j \in \mathcal{N}_h(i)$ in a neighborhood (window) of size $ h \times h $, expressed by  
\begin{align}
w_{ij} = \exp{\left\{-\frac{\|\x_i-\x_j\|_2^2}{2\sigma^2}\right\}}, \text{    } \forall j\in \mathcal{N}_h(i). \nonumber
\end{align}
The parameter $ \sigma $ controls the acceptable change in illumination due to noise, compression artifacts, etc.
Notice that \mbox{$ w_{ij} \rightarrow 1 $} indicates that the patch $ \x_i $ and $ \x_j $ are highly similar, while \mbox{$ w_{ij} \rightarrow 0 $} indicates that the two are substantially different. The size of the surrounding window can be the whole image, or a smaller portion of it. Intuitively, when the majority of $ w_{ij} $ are high, we can conclude that the small central patch $ \x_i $ has many co-occurrences in its surroundings, e.g., it might originate from a large flat area. Similarly, if most of the weights are small, the patch $ \x_i $ is unique, e.g., it might originate from a highly textured and non-repetitive area. As such, these weights provide information about the geometric structure of a large patch w.r.t. its small central patch. 

Next, in order to have a compact representation, we suggest a rearrangement of the correlation surface $ \{w_{ij}\}_{j\in \mathcal{N}_h(i)} $ of the patch $ \x_i $ into a histogram of $ b $-bins (e.g. $ b=10 $). Finally, histogram normalization step is applied, leading to the context-feature $ \H(\x_i) $, which is an empirical distribution of the co-occurrences of the central patch in its large surroundings.    
Put differently, $ \H(\x_i) $ measures the correlations between the central patch to its surroundings: When the correlation is relatively high (many co-occurrences) the histogram is biased towards $ \delta(b_k-10) $, and when the correlation is low the histogram concentrates around $ \delta(b_k-1) $, where $ b_k $ is the $ k $-th bin of the histogram.

A visualization of the context-feature is given in Fig. \ref{context_visualization}. We compute the proposed feature $ \H(\x_i) $ for each patch $ \x_i $ of \textsf{Surfer} image. The size of the small patch and its surrounding window are $ 7 \times 7 $ and $ 21 \times 21 $, respectively. The similarity weights $ \{w_{ij}\}_{j\in \mathcal{N}_h(i)} $ are partitioned into histogram of 10 bins, where the different weights are evenly spaced (the space is 0.1)\footnote{A detailed discussion about the effect of the length of the histogram is given hereafter.}. An illustration of the normalized histogram of 3 different patches, taken from \textsf{Surfer} image, is given in Fig. \ref{context_visualization}b. The green patch in Fig. \ref{context_visualization}a has a small amount of co-occurrences in its large surroundings, thus most of its similarity weights are small, leading to the green curve (with square marker) that concentrates around the first bin (weights between 0 to 0.1). On the other hand, the blue patch in Fig. \ref{context_visualization}a is highly correlated to its surrounding patches, leading to the blue curve (with diamond marker), which concentrates around the last two bins (weights between 0.8 to 1). The red patch has "medium" amount of co-occurrences, thus the red curve in Fig. \ref{context_visualization}b is spread over wide range of bins.

The 10 heat-maps in Fig. \ref{context_visualization}c -- Fig. \ref{context_visualization}l plot the values of the features, computed per each patch, where warmer color corresponds to larger value. Notice that the $ i $-th pixel in each heat-map corresponds to a specific bin of the histogram, representing one element of $ \H(\x_i) $. The $ i $-th pixel along these 10 heat-maps constructs the whole feature $ \H(\x_i) $, i.e., the 1D normalized histogram. Fig. \ref{context_visualization}c plots the values of the first bin in $ \H(\x_i) $ for all the patches, while Fig. \ref{context_visualization}l plots the values of the last bin. As can be seen, warm pixels in Fig. \ref{context_visualization}c correspond to patches that are different than their surroundings (e.g. the foam of the wave and the surfer's face), while warm pixels in the latter heat-map correspond to patches that are similar to their neighbors (e.g. the sky which is a flat area). In some sense, this feature is a measure of the complexity of the patch, ''simple'' patches tend to be correlated with their neighbors, while complex patches are unique and has a few (if any) nearest neighbors.

\begin{algorithm}
	\caption{Evaluating the Con-Patches}
	\label{alg1}
	\begin{algorithmic}[1]
		\REQUIRE $\x\in\RR^{N}$ -- Input image; \\
		\qquad \ \ $\sigma$ -- Acceptable change in illumination; \\
		\qquad \ \ $h$ -- Context window width/height; \\
		\qquad \ \ $b$ -- Histogram length; \\
		\qquad \ \ $\alpha$ -- Context gain; \\
		\qquad \ \ $c$ -- Patch width/height.
		
		\FOR { $ i = 1 \ \TO \ N $ }
		\FOR { $ j \in \mathcal{N}_h(i) $ }
		\STATE $ w_{ij} = \exp{\left\{-\frac{\|\x_i-\x_j\|_2^2}{2\sigma^2}\right\}} $, where $ \x_i,\x_j\in\RR^{c^2} $ are patches of size $ c \times c $.
		\ENDFOR
		\STATE $ \H(\x_i) \Leftarrow $ Partition $ \{w_{ij} \}_{j \in N_h(i)} $ weights into a histogram of $ b $ bins, followed by normalization to unit sum.
		\STATE $ \x_i^{\text{con}} = \left[ \begin{array}{c} \x_i \\ \sqrt{\alpha}\text{ }\H(\x_i) \end{array} \right]$.
		\ENDFOR
		\RETURN $ \x_i^{\text{con}}, \ i=1,...,N $.
	\end{algorithmic}
\end{algorithm}

In the case of external image denoising, Fig. \ref{set_bins} demonstrates the effect of the length of the histogram on the denoising performance for various noise levels, where the con-patches are constructed according to Algorithm \ref{alg1}. More details about external denoising can be found in Section \ref{denoising}. Also, a thorough explanation about the construction of the con-patches is given in the next subsection. The restoration is evaluated by averaging the Peak Signal to Noise Ratio (PSNR) over 10 images, randomly chosen from BSDS300 dataset \cite{BSDS300}, where the higher the PSNR the better the restoration. The similarity weights $ w_{ij} $ are partitioned into histogram of $ b $ equally-spaced bins ($ b \in \{0, 2, 4, 8, 16\}$), followed by a normalization step. The importance of the context-feature (a scalar, denoted by $ \alpha $, which multiplies the histogram) is optimally tuned per each histogram-length and noise-level. As can be seen, increasing the number of bins from $ b=0 $ (i.e., use the conventional patch) to $ b=8 $ consistently improves the restoration, demonstrating the effectiveness of the context. Using a finer grid than 8 bins does not gain additional improvement, indicating the compactness of the proposed feature.



\subsection{Constructing the Con-Patches}
\label{context_patch}

In the following discussion we assume that the distance between two patches is defined as
\begin{align}
\label{dij}
d^2(\q,\p_j) = {\|\q-\p_j\|_2^2},
\end{align}
where $ \q\in\RR^{h\times h} $ is an input query patch and $ \p_j\in\RR^{h \times h} $ is a sample from the database. Both $ \q $ and $ \p_j $ are \emph{large patches} of size \mbox{$ h\times h $}, held as column vectors after lexicographic ordering. The rational behind the con-patches is to leverage the additional force that large patches bring but still work with small patches. As such, we aim to replace a portion from the large patch with its compact representation (context-feature). More specifically, using norm properties we can rewrite \mbox{Eq. (\ref{dij})} as
\begin{align}
\label{split_dij}
d^2(\q,\p_j) = \|\R_c\q-\R_c\p_j\|_2^2 + \|\R_s\q-\R_s\p_j\|_2^2,
\end{align}
where $ \R_c\in\RR^{c^2\times h^2} $ and $ \R_s\in\RR^{s^2\times h^2} $ are operators that extract the central and surrounding parts of a patch, respectively. The central part is the $ c\times c $ small patch, while the surrounding is of size $ s \times s $ pixels ($ c \ll s $). Notice that in our terminology, $ \R_c\q $ is the regular patch, while $ \R_s\q $ is its context, \emph{but in high dimensions}. 

We can dramatically reduce the size of the large patch by replacing the surrounding of the central patch with its context-feature. As such, the modified distance between two patches is expressed by
\begin{align}
\label{low_split_dij}
d_{\H}^{2}(\q,\p_j) &= \|\R_c\q-\R_c\p_j\|_2^2 + \alpha\|\H(\q)-\H(\p_j)\|_2^2 \\ \nonumber
&= \left\| \left[ \begin{array}{c} \R_c\q \\ \sqrt{\alpha}\text{ }\H(\q) \end{array} \right] - \left[ \begin{array}{c} \R_c\p_j \\ \sqrt{\alpha}\text{ }\H(\p_j) \end{array} \right] \right\|_2^2,
\end{align}
where $ \alpha $ is a scalar (gain) that controls the importance/weight of the context. Notice that the con-patches are obtained by concatenating the context-feature to the regular patch. In addition, the size of the con-patch is much smaller than the large patch, while being slightly bigger than the central one. Algorithm \ref{alg1} illustrates the process of constructing the con-patches. Notice that $ \x_i$, $\x_j $ in this algorithm are patches of small size.

\subsection{Testing the Power of Con-Patches}
\label{power_context_patch}

\begin{figure}[!t]
	\centering
	\includegraphics[width=3.2in]{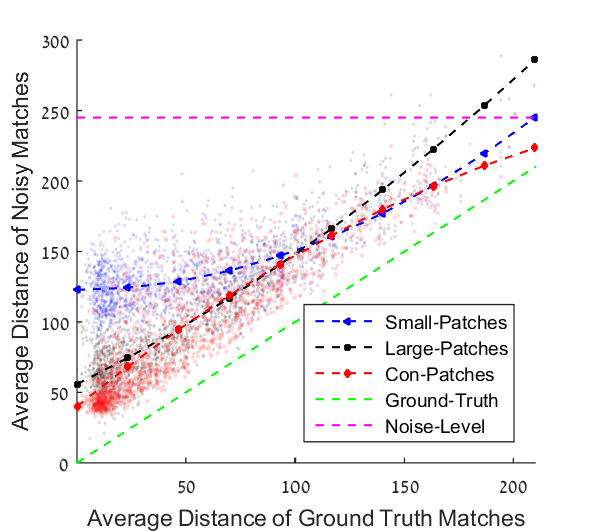}
	\caption{Comparison between applying $ k $-NN using small (blue triangles), large (black squares) and con-patches (red diamonds). Refer to Section \ref{power_context_patch} for a detailed explanation.}
	\label{scatter}
\end{figure}

The above description may suggest that the true distance between the large patches and the approximated one obtained via the con-patches tend to be similar. In this section we address this interpretation and show that while this is not exactly the case, a related and highly desired property does hold true.

The following experiment illustrates the superiority of applying $ k $-NN search using the proposed context-based distance $ d_{\H}^{2}(\q_i,\p_j) $ over both the conventional distance between large $ d^{2}(\q_i,\p_j) $ or small $ d^{2}(\R_c\q_i,\R_c\p_j) $ patches, all in the context of noisy data. In what follows we propose an experiment in which we show that the con-patches lead to more appropriate matches, when compared to the above two alternatives. 

The size of a small patch is $ 7 \times 7 $, which is the center part of a large patch of size $ 17 \times 17 $. The con-patch is composed of the small central patch (of size $ 7\times 7 $), concatenated to an 8 bin context-feature that represents its $ 17\times 17 $ surroundings. 

We gather two sets of clean large image patches of size $ 17 \times 17 $ pixels: query-set $ \{\q_i\}_{i=1}^{N_T} $ and example-set $ \{\p_j\}_{j=1}^{N_D} $, where $N_T =  8\times 10^5 $ and $ N_D = 35\times 10^5 $. In addition, a noisy version of the query-set, denoted by $ \{\y_i\}_{i=1}^{N_T} $, is created by contaminating the patches with zero-mean Gaussian noise with $ \sigma_v = 35$. All these patches are randomly chosen from BSDS300 database\cite{BSDS300}, where we ensure that the noisy query-set patches $ \{\y_i\}_{i=1}^{N_T} $ have at least $ k=20 $ NNs in the example-set $ \{\p_j\}_{j=1}^{N_D} $ (in the sense of having an Euclidean distance that falls below the noise level).

A set of ground-truth matches per each query-patch is created by applying $ k $-NN search in the example-set based on the \emph{small central part (of size $ 7\times 7 $) of the clean query-patch} $ \R_c\q_i $. Then, we evaluate the average $ l_2 $ distance between $ \R_c\q_i $ and its $ k=20 $ ground-truth matches, denoting this distance by $ \E_{\text{GT}}(i) $. Since this computation is done using the clean patches, these stand as the ultimate distances between each small query patch and the candidate neighbors from the example-set. 

Next, $ k $-NN search is applied for each \emph{noisy query-patch}, using the 3 variants of the distances: 
\begin{enumerate}
	\item Using small patches of size $ 7\times 7 $ $, d^{2}(\R_c\y_i,\R_c\p_j)  $,
	\item Using large patches of size $ 17\times 17 $, $ d^{2}(\y_i,\p_j) $, and
	\item Using the con-patches, with the whole center part of size $ 7\times 7 $ and an augmented feature of length 8 to represent the surrounding of size $ 17\times 17 $, $ d_{\H}^{2}(\y_i,\p_j) $,
\end{enumerate}
leading to $ \E_{\text{small}}(i) $, $ \E_{\text{large}}(i) $ and $ \E_{\text{con}}(i) $, respectively. Notice that the matching process is done based on the \emph{noisy} patches, but the average error/distance $ \E $ is always computed based on the small central part of the clean patches. As such, the closer the average distance to the average ground-truth distance $ \E_{\text{GT}}(i) $, the better the matching process. 

Fig. \ref{scatter} illustrates these average distances between 1500 randomly chosen query-patches and their matches. The horizontal coordinate of each point $ i $ in the scatter-plot represents the value of $ \E_{\text{GT}}(i) $, while the vertical coordinates represent the values of $ \E_{\text{small}}(i) $, $ \E_{\text{large}}(i) $, and $ \E_{\text{con}}(i) $, respectively. In addition, the dashed blue-triangle, black-square, and red-diamond curves correspond to polynomial fits (of degree 2) of $ \E_{\text{small}} $, $ \E_{\text{large}} $, and $ \E_{\text{con}} $, respectively, where the fits are computed based on the whole query-set ($ 8\times 10^5 $ patches). The dashed green line represents the case in which the horizontal and vertical coordinates are equal, thereby fulfilling it implies that all the right neighbors are found. The dashed magenta line reflects the noise-level and it is equal to $ n\cdot \sigma = 7\cdot 35 $, i.e., the distances that are below this line are valuable, being smaller than the noise-level.

Following Fig. \ref{scatter}, when the test-set patches have many meaningful nearby samples in the example-set (in the region that \mbox{$ \E_{\text{GT}} < 100 $}), the curves of $ \E_{\text{large}} $ and $ \E_{\text{con}} $ are comparable and much better than $ \E_{\text{small}} $. In this region the "dense-sampling" assumption holds, and applying $ k $-NN search using large or con-patches results in equally good set of neighbors, while working with small patches tends to err, leading to "pseudo-matches". In this regime the proposed con-patches successfully seize the important information that large patches provide, in a very compact way while being robust to the noise. 

On the other hand, for relatively large ground-truth distances ($ \E_{\text{GT}} > 100 $), the $ \E_{\text{con}} $ curve is found to be the best\footnote{Notice that for $ 110 < \E_{\text{GT}} < 160 $ the small-patches perform slightly better than con-patches. This possibly can be solved by setting $ \alpha $ in an adaptive fashion, which can be considered as a promising future research direction.}. In this regime there is a limited amount of nearby samples, and thus working with large patches is inferior to both small or con-patches.

This experiment shows that the context proposed is beneficial across the board -- it leads to better matches than the ones obtained by small or large patches. This experiment also demonstrates that our context-feature exploits the unique force of large patches, along with the attractive "dense-sampling" property of small ones. 

To conclude, the proposed con-patch benefits from the simplicity and convenience of working with small patches, as it adds a short feature vector to each patch. On the other hand, con-patches bring an additional force that relates to large patches, and does so without the need to have a denser sampling of the patch manifold. We have seen that beyond these properties, con-patches introduce a robustness to noise which enables them to perform even better that large patches. 
In the following section we demonstrate the effectiveness of the con-patches on various applications.

\section{Experimental Results}
\label{experiments}

In this section we test the proposed idea on three problems: (i) External image denoising, (ii) Depth single image SR, and (iii) MC-FRUC. The core restoration mechanism in these problems is a NN search, while each problem has different degradation model: additive noise, missing of spatial or temporal content, respectively. We show that a significant boost in performance is achieved by simply plugging the context-feature to the conventional patches.

\subsection{External Denoising}
\label{denoising}

The image denoising problem \cite{buades2005non,elad2006image,zontak2011internal, zoran2011learning,levin2011natural,levin2012patch,romano2013improving, mosseri2013combining,romano_disagree2015,adaptive2015database,ghimpeteanu2016decomposition,papyan2016multi} assumes that a noisy measurement $ \Y\in \RR^{N} $ of a clean image $ \Z\in \RR^{N} $ is obtained by a degradation of the form
\begin{align}
\Y = \Z + \V, 
\end{align} 
where $ \V\in\RR^N $ is an i.i.d. zero-mean Gaussian noise with standard-deviation $ \sigma_v $. Our goal is to restore the unknown underlying image $ \Z $ from $ \Y $ using a large database of clean images. More formally, external denoising methods break the noisy image $ \Y $ into overlapping patches $ \y_i $ and denoise each patch by applying a weighted average over the $ k $-NN of the patch $ \y_i $, expressed by
\begin{align}
\label{denoise_patch}
\xh_i = \frac{1}{\sum_{j=1}^{k}{\exp\left\{-\frac{\|\y_i-\x_i^j\|_2^2}{2\sigma_{v}^2}\right\}}}\sum_{j=1}^{k}{\exp\left\{-\frac{\|\y_i-\x_i^j\|_2^2}{2\sigma_{v}^2}\right\}\x_i^j},
\end{align}
where $ \xh_i $ is the denoised version of the noisy patch $ \y_i $, and $ \{\x_i^j\}_{j=1}^{k} $ are the nearby patches that are found in the database. Then, the denoised image $ \xh $ is constructed from the locally cleaned patches by returning each patch $ \xh_i $ to its  \mbox{$ i $-th} location in the global image, followed by averaging over the overlaps.

\begin{figure}[!t]
	\centering
	\subfloat[$ \sigma_v \in \{15, 25, 35\} $]{\includegraphics[width=1.7in, trim = 0 0 0 0]{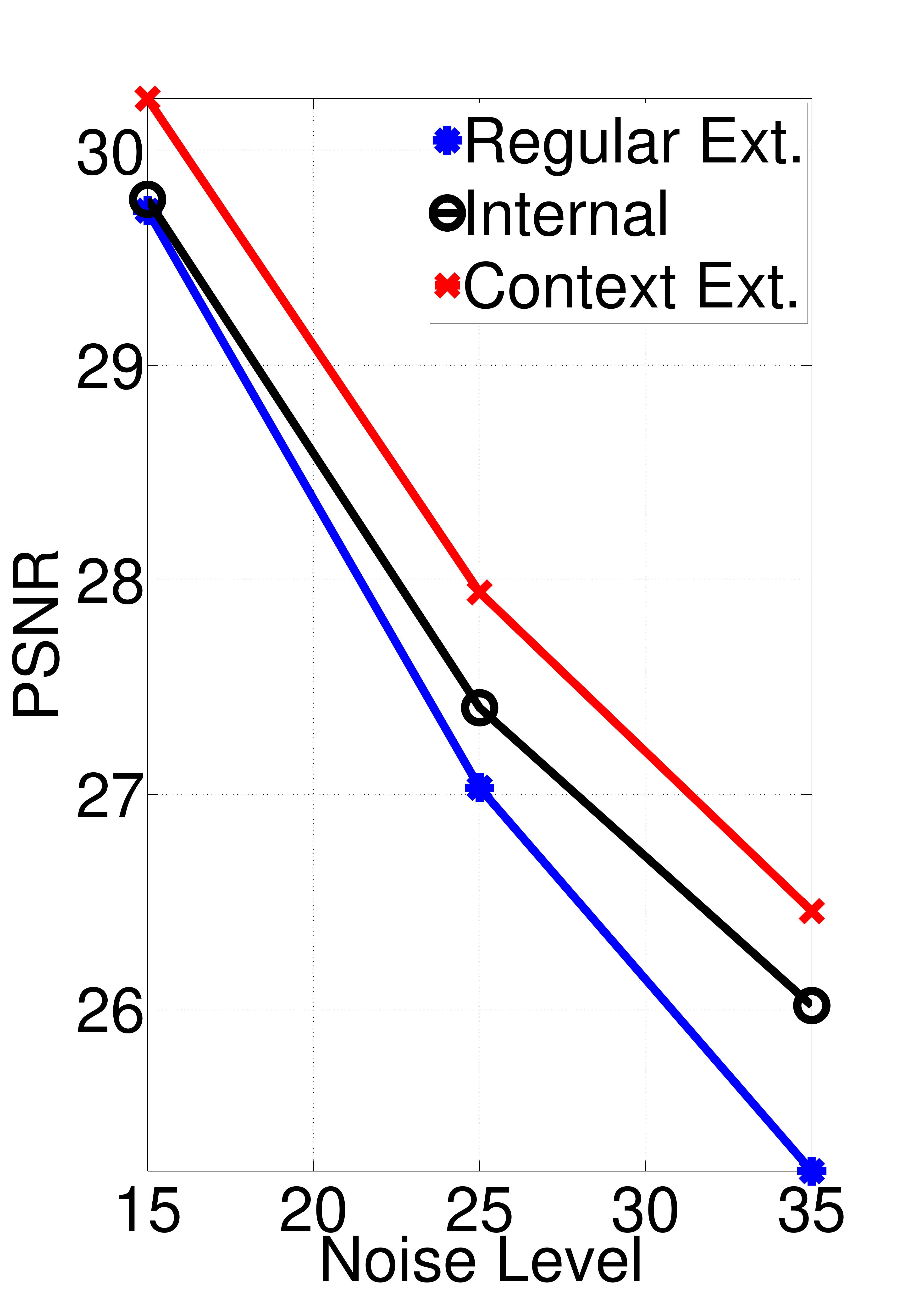}}
	\centering
	\subfloat[$ \sigma_v \in \{50, 75, 100\} $]{\includegraphics[trim= 0 0 0 150, width=1.7in]{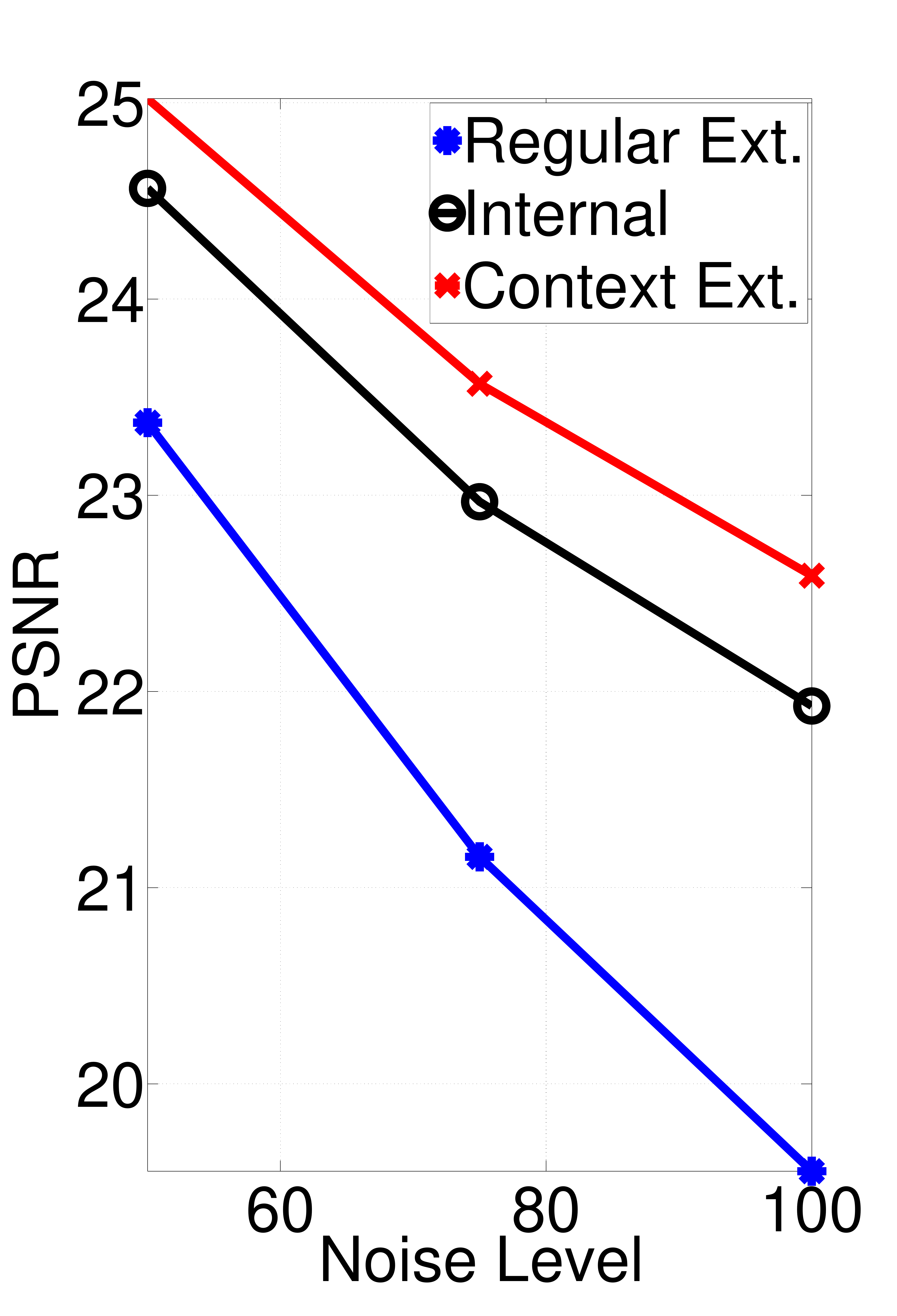}}
	
	\centering
	\caption{Denoising results (in terms of PSNR) of conventional external denoising, internal NLM, and the proposed context-based external denoising, averaged over 100 test-set images of BSDS300 database. (a) Comparison for noise levels 15, 25, 35, and (b) Comparison for noise levels 50, 75, 100.}
	\label{denoising_res}
\end{figure}

\begin{figure*}[ht]
	\centering
	\subfloat[Noisy (24.62dB)]{\includegraphics[scale=0.28]{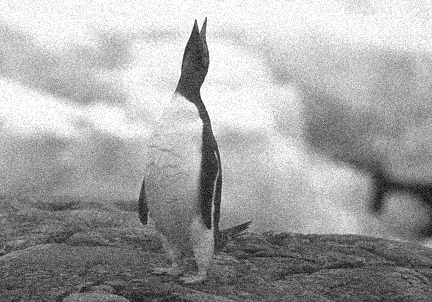}}
	\hfil
	\centering
	\subfloat[Internal (33.54dB)]{\includegraphics[scale=0.28]{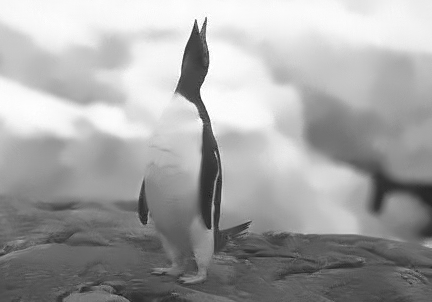}}
	\hfil
	\centering
	\subfloat[Regular-External (32.46dB)]{\includegraphics[scale=0.28]{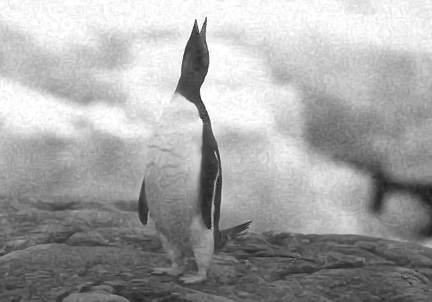}}
	\hfil
	\centering
	\subfloat[Context-External (34.25dB)]{\includegraphics[scale=0.28]{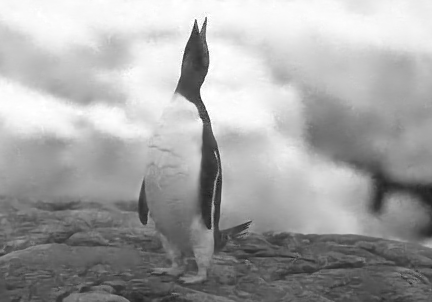}} \\
	
	\centering
	\subfloat[Noisy (24.62dB)]{\includegraphics[scale=0.28]{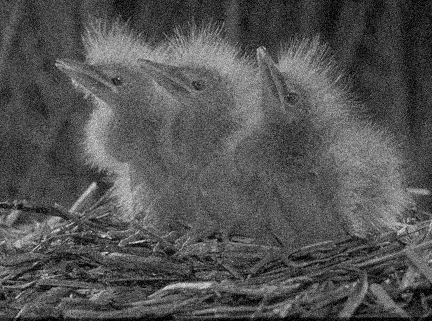}}
	\hfil
	\centering
	\subfloat[Internal (30.89dB)]{\includegraphics[scale=0.28]{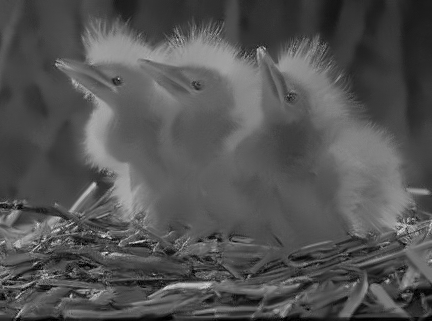}}
	\hfil
	\centering
	\subfloat[Regular-External (30.94dB)]{\includegraphics[scale=0.28]{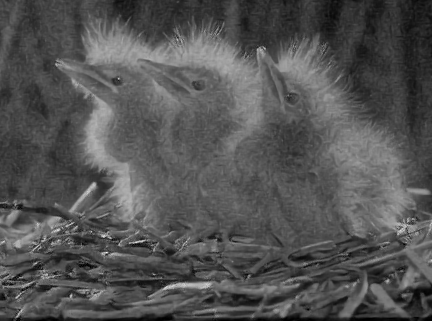}}
	\hfil
	\centering
	\subfloat[Context-External (31.51dB)]{\includegraphics[scale=0.28]{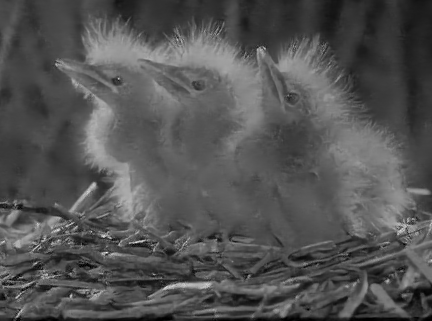}}
	\centering
	
	
	\caption{Visual comparison for noisy ($ \sigma = 15 $) \textsf{Penguin} and \textsf{Baby-Birds} images: (a,e) Noisy images, (b,f) Internal NLM, (c,g) External denoising using conventional patches, and (d,h) External denoising using con-patches.}
	
	\label{visual_compare_denoising}
\end{figure*}

\begin{figure*}[ht]
	\centering
	\subfloat[$ \sigma_v = 15 $]{\includegraphics[width=1.7in]{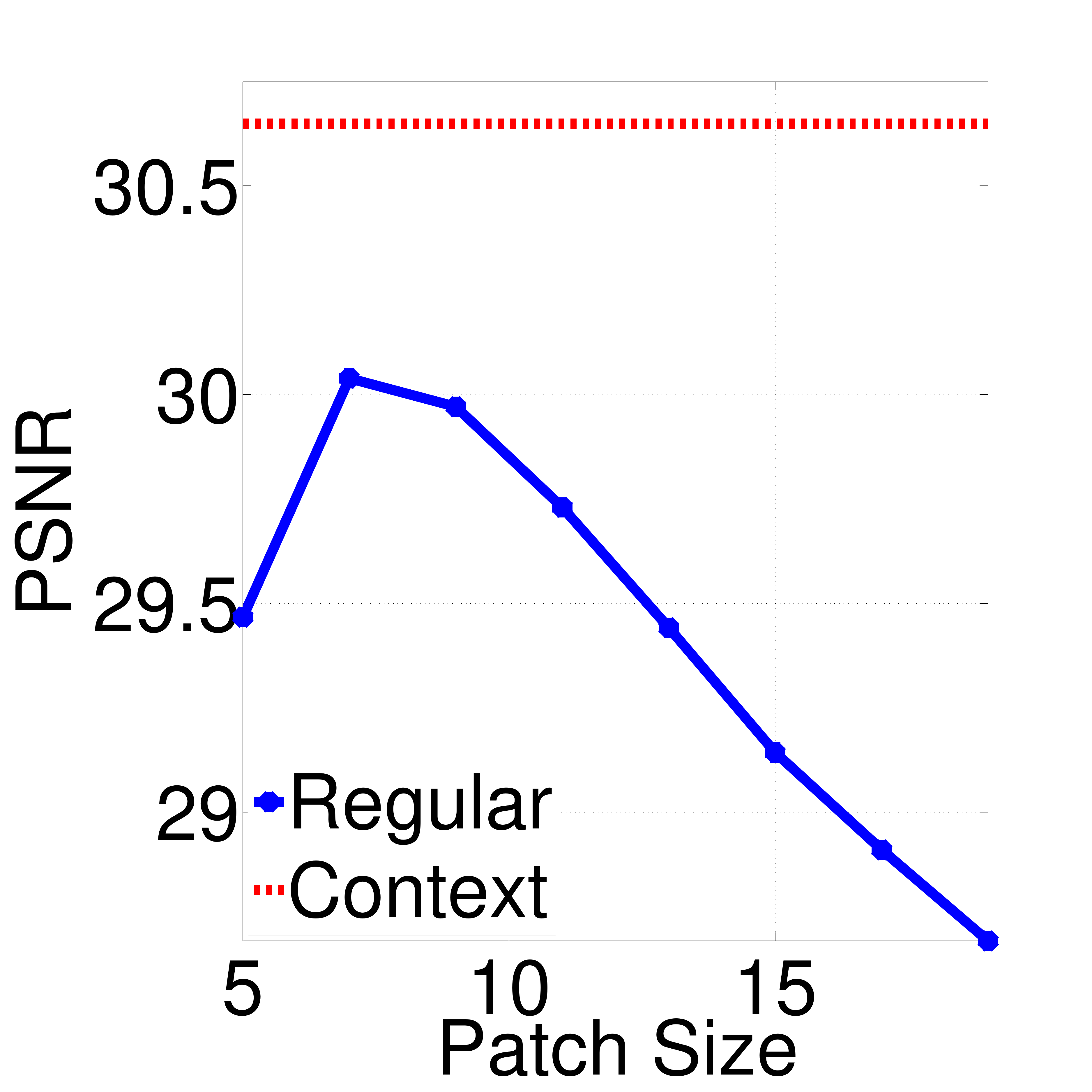}}
	\centering
	\subfloat[$ \sigma_v = 25 $]{\includegraphics[width=1.7in]{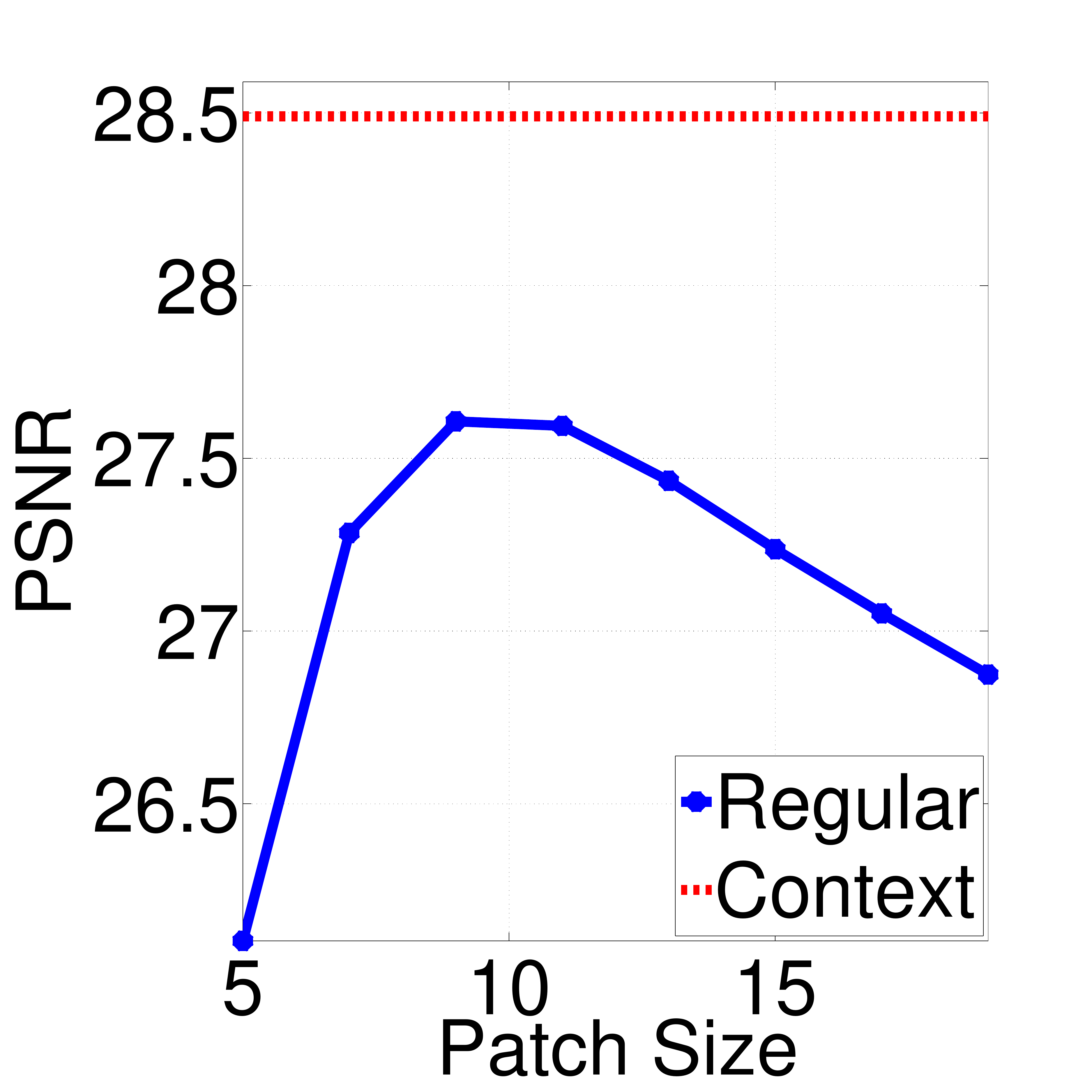}}
	\centering
	\subfloat[$ \sigma_v = 35 $]{\includegraphics[width=1.7in]{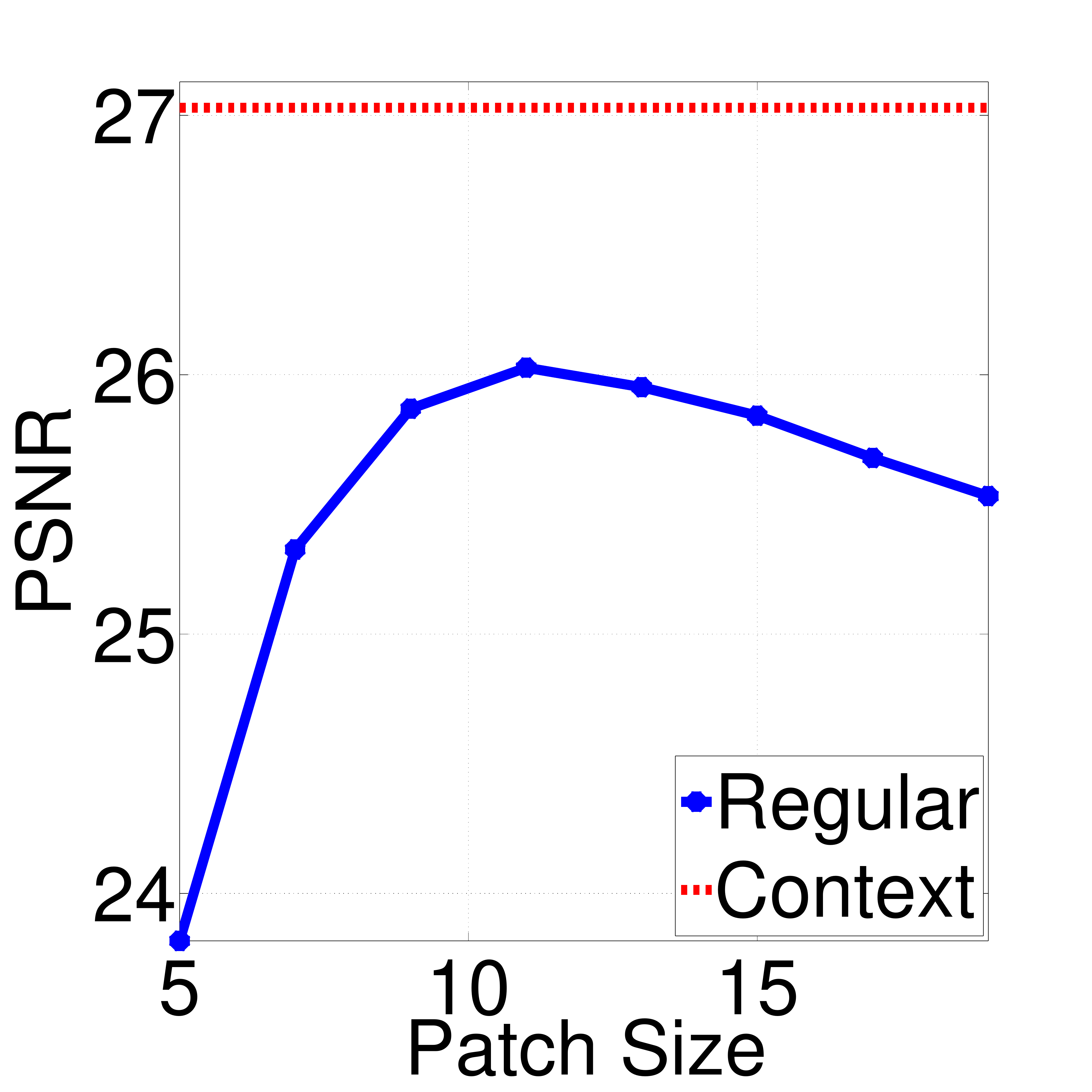}}
	\centering
	\subfloat[$ \sigma_v = 50 $]{\includegraphics[width=1.7in]{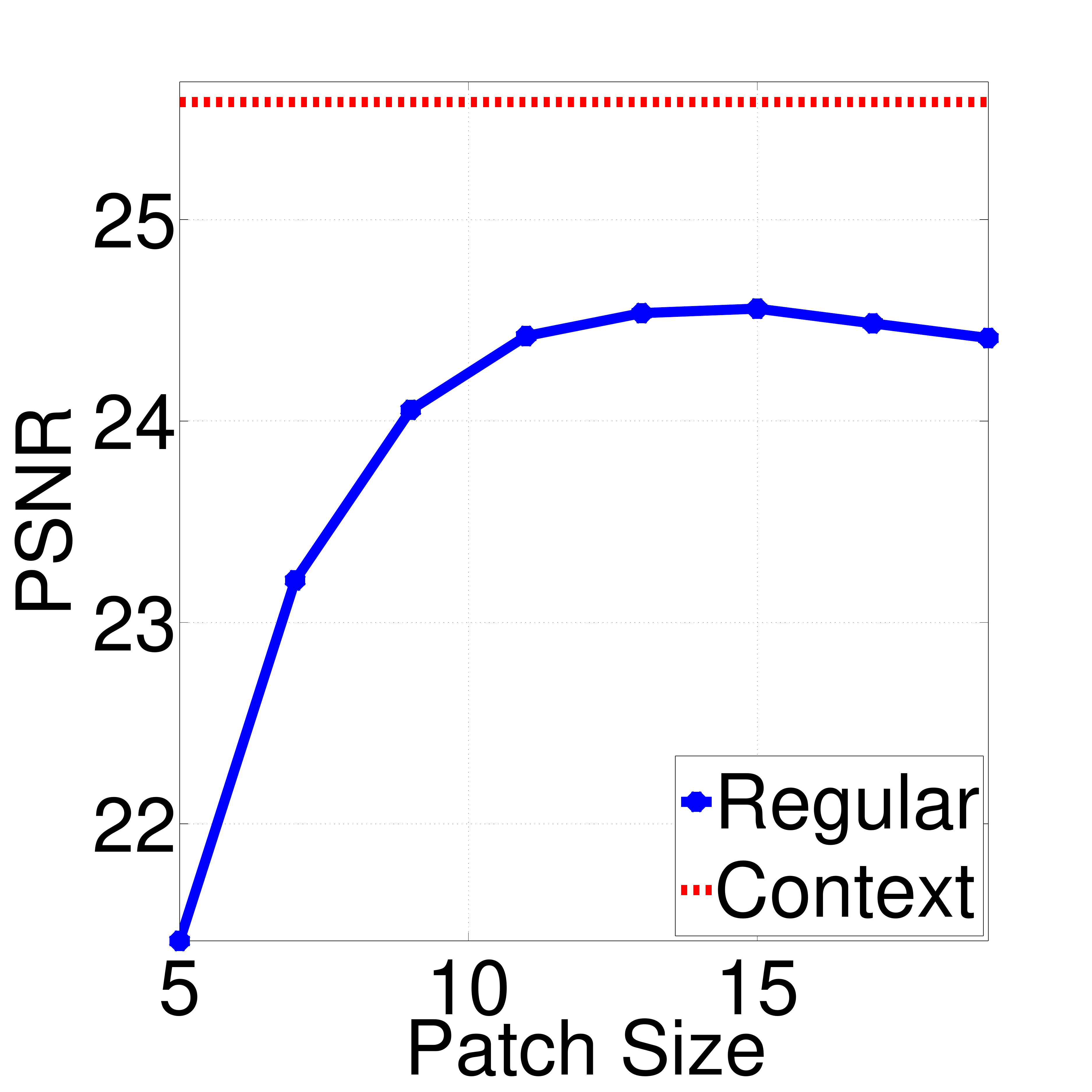}}
	\centering
	\caption{The effect of the patch-size on conventional external denoising performance for different noise-levels (blue solid curves), along with a comparison to the proposed context-based approach (red dashed line). The size of the con-patch is fixed to $ 7\times7 $ patch, concatenated to a feature of length $ b=8 $. We measure the average PSNR over 10 randomly chosen images from the test-set of BSDS300 database, obtained by applying conventional external denoising with patches of varying size, starting from $ 5\times 5 $ up to $ 19\times 19 $. The effect of the patch-size is tested for the following noise levels: (a) $ \sigma_v = 15 $, (b) $ \sigma_v = 25 $, (c) $ \sigma_v = 35 $, and (d) $ \sigma_v = 50 $. }
	\label{var_psz}
\end{figure*}

Generally speaking, given the noisy patches $ \y_i $, the challenge is to find several clean database samples $ \{\x_i^j\}_{j=1}^{k} $ that are close to the unknown patch $ \z_i $. The closer the matched samples to $ \z_i $, the better the restoration. In this experiment we show that improved restorations are achieved when using con-patches in the NN search rather than the conventional patches.

In order to plug the context we should compute the context-feature, both on the clean database images and the input noisy ones. The parameter $ \sigma $ in Eq. (\ref{wij}), which controls the acceptable change in illumination, is set to be a fixed value ($ \sigma=5 $) for the clean database images, and equal to the noise-level $ \sigma_v $ for the noisy images. Per each noise level $ \sigma_v$, we tuned the parameter $ \alpha $, the gain which controls the importance of the context feature, resulting in 
\begin{align}
 \sqrt{\alpha} = 
  \begin{cases} 
   0.9 & \sigma_v < 50 \\
   1.1 & 50 \leq \sigma_v < 100 \\
   1.3 & \sigma_v \geq 100
  \end{cases} \nonumber  
\end{align}
The rest of the parameters are fixed for all noise-levels; the small patch is of size $ c\times c = 7 \times 7 $, the context-feature is of length $ b=8 $, computed on a window of size $ h \times h = 21\times 21 $ with step-size of $ m=4 $ pixels. The denoising is done by searching for the $ k=500 $ NN in an external database of $ 4.5 \times 10^6 $ randomly sampled patches, taken from the training-set images of BSDS300 dataset \cite{BSDS300}. Notice that the results of external denoising using conventional patches are obtained by setting $ \alpha = 0 $ in the NN search.

Note that, in order to reduce computations, we have chosen to use kd-tree as a way to approximate the $ k $-NN search. This results in query time of roughly $ \mathcal{O}(L+k) $, where $ L $ is the size of the database, leading to runtime of about 2 minutes for denoising a grayscale image of size $ 480\times320 $ (on a laptop with Intel Core i7-3940XM, 3.00GHz CPU). In this context we should note that internal methods tend to be faster than the external ones. For example, the runtime of the BM3D \cite{dabov2007image} is about 2 seconds of image of this size (evaluated on the same computer).
	
The proposed algorithm is evaluated based on the 100 test-set images of BSDS300 database. Fig. \ref{denoising_res} demonstrates the denoising results for various noise levels in terms of PSNR, where we compare the context-based external denoiser to its conventional version \cite{levin2011natural,levin2012patch,zontak2011internal,mosseri2013combining}, and also to internal NLM \cite{NL_DENOISE_REF4}. As demonstrated in \mbox{Fig. \ref{denoising_res}}, external denoising using con-patches is significantly better than the conventional patches, indicating that the context has a major impact on the patch matching mechanism. Quantitatively, on average, for noise levels of 15, 25, 35, 50, 75, and 100 plugging the context leads to improvement of 0.52dB, 0.91dB, 1.21dB, 1.65dB, 2.41dB, and 3.03dB, respectively.

\begin{table*}[htbp]
	\centering
	\caption{Comparison between the results [PSNR] of the conventional/regular external SR \cite{mac2012patch} and the proposed context-based variant. The best results per each image and scaling factor are highlighted.}
	\begin{tabular}{||C{2.3cm}||C{1.2cm}|C{1.2cm}|C{1.2cm}|C{1.2cm}|C{1.2cm}|C{1.2cm}|C{1.2cm}||C{1.2cm}||C{1.2cm}||}
		\hline
		\textbf{Method (Scaling)} & \textsf{Cones}    & \textsf{Teddy}    & \textsf{Tsukuba}  & \textsf{Venus}    & \textsf{Scan 21}  & \textsf{Scan 30}  & \textsf{Scan 42}  & \multicolumn{1}{c||}{\textbf{Average}} \\
		\hline\hline
		{Regular (x2)} & 32.51  & 30.64  & 20.62  & 33.78  & 41.80  & 42.59  & 37.97  & 34.27 \\ \hline
		{Context (x2)} & \textbf{33.11} & \textbf{31.57} & \textbf{21.55} & \textbf{34.50} & \textbf{42.05} & \textbf{42.88} & \textbf{38.49} & \textbf{34.88} \\ \hline \hline
		{Regular (x4)} & 29.15  & 28.56  & 19.16  & 30.22  & 39.62  & 39.74  & 32.55  & 31.29 \\ \hline
		{Context (x4)} & \textbf{29.98} & \textbf{29.30} & \textbf{19.33} & \textbf{32.68} & \textbf{40.18} & \textbf{40.85} & \textbf{34.33} & \textbf{32.38} \\ \hline
	\end{tabular}%
	\label{depth_compare}%
\end{table*}%

\begin{figure*}[ht]
	\centering
	\subfloat[Original]{\includegraphics[scale=0.32]{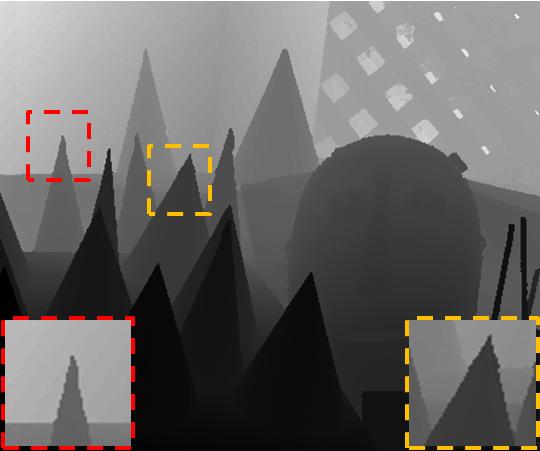}}
	\hfil
	\centering
	\subfloat[LR]{\includegraphics[scale=0.32]{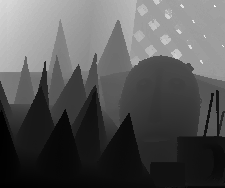}}
	\hfil
	\centering
	\subfloat[Regular-External]{\includegraphics[scale=0.32]{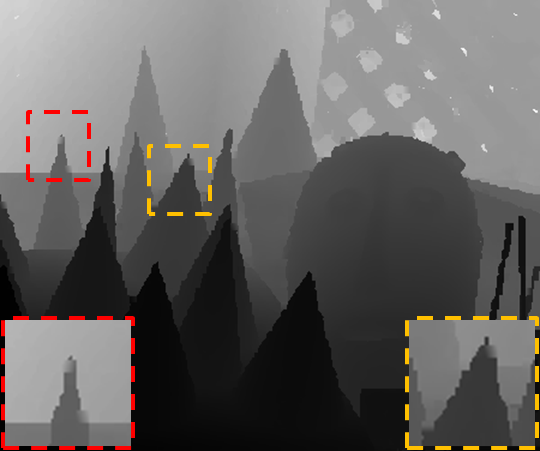}}
	\hfil
	\centering
	\subfloat[Context-External]{\includegraphics[scale=0.32]{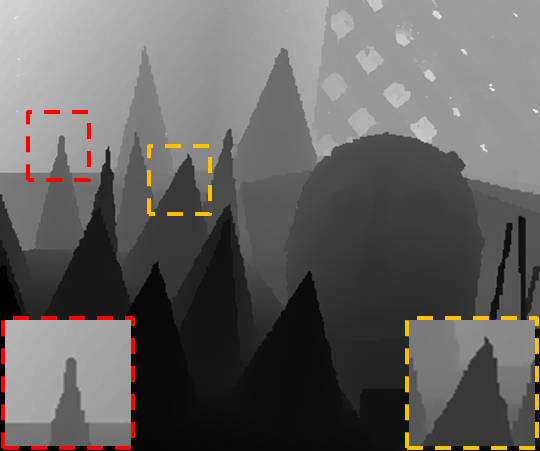}}
	\centering
	\caption{Visual comparison for upscaling \textsf{Cones} depth image by a factor of 2. (a) Original image, (b) LR input image, (c) External SR using conventional patches \cite{mac2012patch} -- 32.51dB, and (d) External SR using con-patches -- 33.11dB.}
	\label{SR_results}
\end{figure*}

A visual comparison between external denoising with and without integrating the context is given in Fig. \ref{visual_compare_denoising}. As can be seen, con-patches lead to a pleasant restoration of the underlying image, both in flat (e.g. the skies) and structured areas (e.g. the rocks or twigs). On the other hand, conventional patches result in an effective restoration of structured/active areas, but fail to recover the flat ones. This observation stands in agreement with \cite{zontak2011internal,mosseri2013combining}, which demonstrate that conventional external denoising tends to over-fit the noise details. Put differently, it has been shown \cite{mosseri2013combining} that random-noise patches have high correlation with natural image patches, explaining why the conventional external denoiser introduces undesired structures, especially in flat ares.

Turning to internal NLM, the results in Fig. \ref{denoising_res} are obtained by fine-tunning the NLM \cite{NL_DENOISE_REF4} parameters to produce the best PSNR for patches of size $ 7\times 7 $ and search window of size $ 21 \times 21 $. Note that we do not test the proposed con-patches on internal NLM since it already seeks for NNs in a limited window, which can be considered as the internal context of the patch. As demonstrated, internal NLM outperforms its conventional external variant, complying with the conclusions of \cite{zontak2011internal}. However, this observation is no longer valid when integrating the context of the patch. As can be seen, external denoising with con-patches outperforms the internal version with average gaps of 0.47dB, 0.54dB, 0.44dB, 0.46dB, 0.60dB, and 0.66dB, for noise-levels 15, 25, 35, 50, 75, and 100, respectively. The achieved gain in performance is also reflected visually, as demonstrated in Fig. \ref{visual_compare_denoising} (notice the improved restoration in the rocks area of \textsf{Penguin} image and the twigs area of \textsf{Baby-Birds} image).

Fig. \ref{var_psz} demonstrates that the proposed con-patch approach benefits from the advantages of working with both large and small patches. It plots the restoration performance of external denoising using the proposed con-patches (flat red dashed-line), compared to conventional patches of varying size (blue curves) -- starting from $ 5 \times 5 $ ($ c = 5 $) up to $ 19 \times 19 $ ($ c = 19 $). In this experiment the \emph{size of the con-patch is fixed}, obtained by concatenating $ 7 \times 7 $ patch to context-feature of length 8. Broadly speaking, as seen in Fig. \ref{var_psz}, the PSNR increases with the conventional patch-size due to improved matching, up to a point where the PSNR starts decreasing due to the lack in valid NN. Note that this observation fits the theoretical conclusions of \cite{levin2011natural,levin2012patch}. However, the context-based approach leads to restoration that outperforms the conventional approach (for all patch-size), demonstrating the power of the context, which benefits from improved patch matching while keeping the "dense-sampling" property.

Notice that the goal of this work is to emphasize the importance of the context. As a way to keep the simplicity and intuition for the source of improvement, we choose the external NLM algorithm as the core denoising mechanism. We should note that our results are inferior to the state-of-the-art denoising methods, such as the BM3D \cite{dabov2007image} and WNNM \cite{gu2014weighted}. More specifically, based on the 100 test images, the BM3D outperforms the external context-based approach in about 0.5dB on average over the tested noise-levels. Similar conclusions hold for the state-of-the-art WNNM. However, notice that based on the image denoising bound work \cite{levin2012patch}, the size of the database is the key for achieving the MMSE estimator. As such, the achieved results can possibly be improved by increasing the size of the database, from $ 4.5\times10^6 $ patches to $ 10^{10} $ (as done in \cite{levin2012patch}), but we choose to avoid this path of research at this point.

To summarize, we illustrated that a significant improvement in denoising performance is achieved simply by plugging the context-feature to the conventional patch in the NN search. We also show how context-based external denoising outperforms the conventional internal version. Finally, by comparing the performance of the context-based denoising to the one obtained by conventional large patches, we show that working with con-patches is not equivalent and even might be much better than the direct use of large patches. 

\begin{table*}[!htbp]
	\centering
	\caption{Comparison between the results [PSNR] of the conventional block-based MC-FRUC \cite{mac2012patch} and its context-based variant, for temporal up-sampling by a factor of 2. The best results per each video sequence are highlighted.}
	\begin{tabular}{||C{1.2cm}||C{1.2cm}|C{1.2cm}|C{1.2cm}|C{1.2cm}|C{1.2cm}|C{1.2cm}||C{1.2cm}||C{1.2cm}||}
		\hline
		\textbf{Method} & \textsf{Foreman}    & \textsf{Flower}    & \textsf{Tennis}  & \textsf{Harbour}  & \textsf{Park-Run}  & \textsf{Ice}  &  \multicolumn{1}{c||}{\textbf{Average}} \\
		\hline\hline
		{Regular} & 33.34  & 29.82  & 28.16  & 30.40  & 28.07 & 33.70  & 30.58 \\ \hline
		{Context} & \textbf{33.47} & \textbf{29.95} & \textbf{28.31} & \textbf{30.46} & \textbf{28.13} & \textbf{33.82}  & \textbf{30.69} \\ \hline
	\end{tabular}%
	\label{fruc_compare}%
\end{table*}%

\begin{figure*}[!ht]
	\centering
	\subfloat[Existing Frame at $ (t-1) $]{\includegraphics[scale=0.38]{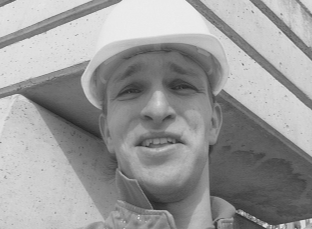}}
	\hfil
	\centering
	\subfloat[Original unknown Frame at $ (t) $]{\includegraphics[scale=0.38]{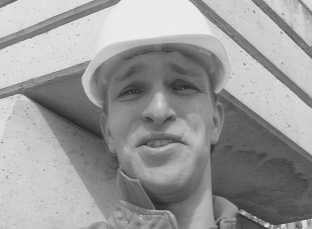}}
	\hfil
	\centering
	\subfloat[Existing Frame at $ (t+1) $]{\includegraphics[scale=0.38]{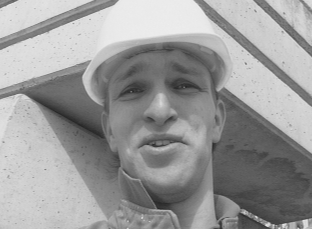}}
	\hfil
	\centering
	\subfloat[Frame Averaging]{\includegraphics[scale=0.38]{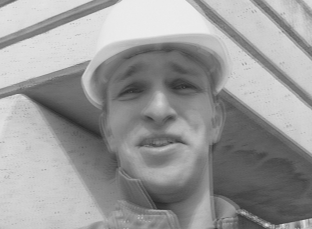}} \\
	\subfloat[Regular]{\includegraphics[scale=0.38]{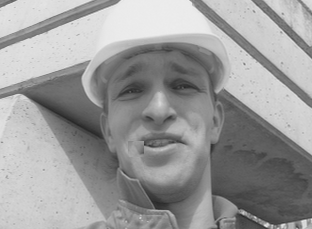}}
	\hfil
	\centering
	\subfloat[Context]{\includegraphics[scale=0.38]{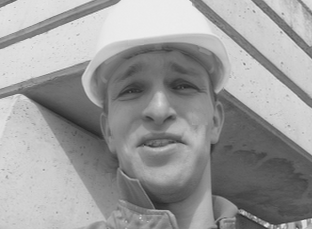}}
	\hfil
	\centering
	\subfloat[Regular Error]{\includegraphics[scale=0.38]{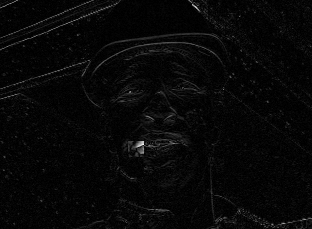}}
	\hfil
	\centering
	\subfloat[Context Error]{\includegraphics[scale=0.38]{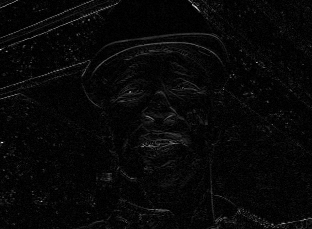}}
	
	\centering
	\caption{Visual comparison for temporal up-sampling \textsf{Foreman} sequence by a factor of 2. (a) Existing frame at time $ t-1 $, (b) Unknown frame at time $ t $ (to be interpolated), (c) Existing frame at time $ t+1 $, (d) Estimation of the $ t $-frame by frame averaging -- 29.42dB, (e) Conventional MC-FRUC -- 35.04dB, (d) Context-based MC-FRUC -- 35.22dB, (d) Absolute value of the error of the conventional approach, and (d) the context-based approach.}
	\label{fruc_results}
\end{figure*}

\subsection{Depth Image Super Resolution}
\label{sr}

Single depth image SR is the process of increasing the resolution of a LR depth image. In this experiment, we build upon the external SR algorithm of Aodha et al. \cite{mac2012patch}, and demonstrate the effectiveness of the proposed con-patches. 

Given a generic database of LR-HR pairs of depth patches, the authors of \cite{mac2012patch} suggest searching for the $ k$-NN of each of the input LR image patches, leading to $ k $ HR candidates per patch. Then, using an MRF model, these candidates are combined together and form the desired HR image. Naturally, the quality of the matches affect the restoration performance. Therefore, we find it interesting to integrate the context-feature to the NN search and compare the results to the conventional approach. In order to have a fair comparison, we follow  the experiments of the external SR work \cite{mac2012patch} and test the two variants of the algorithm on Middlebury stereo dataset \cite{scharstein2002taxonomy}. The LR images are generated by decimating the dataset images by factors of 2 and 4, both in the horizontal and vertical dimensions. The reconstructed images are obtained by applying the original software of \cite{mac2012patch} on the generated LR images. 

The context-feature is computed by measuring the similarity of $ 11 \times 11 $ patch to its $ 21 \times 21 $ neighborhood, where  $ \sigma = 5 $ and $ m=4 $ (both for the dataset and input images). Similarly to image denoising, this feature is plugged \emph{only} in the NN search (with $ \sqrt{\alpha} = 0.4 $) and detached from the patch for the rest of the algorithm. Both in the conventional and its context-based variant, we use the very same parameters that were chosen by the authors.

In Table \ref{depth_compare} we provide a comparison (in terms of PSNR) between the conventional patch-based external SR \cite{mac2012patch} and its context-based variant. As can be seen, the con-patch approach results in a superior restoration performance when compared with the conventional one. Quantitatively, we gain an average improvement of 0.61dB and 1.09dB for upscaling by factors of 2 and 4, respectively. The gain in PSNR is also reflected visually, as demonstrated in Fig. \ref{SR_results} for \textsf{Cones} image, upscaled by a factor of 2. As illustrated, the proposed context-based restoration results in continuous edges, having less artifacts than the conventional version (notice the 2 zoomed-in regions).

To conclude, we test the proposed concept on single depth image SR, showing that the context-feature fits to depth images. This experiment demonstrates how con-patches can effectively boost the restoration of missing spatial content.

\subsection{Motion-Compensated Frame-Rate Up Conversion}
\label{fruc}

Frame Rate Up Conversion (FRUC) is the process of increasing the temporal resolution (frame-rate) of a given video, used for format conversion, improvement of low bit-rate video coding, and more \cite{choi2000new,lee2003weighted,zhai2005low,vinh2009frame,wang2010motion1,wang2010motion2,dong2014adaptive,ydar2014fruc}. Frame averaging or repetition are the simplest FRUC techniques, ignoring the motion between frames, and thereby result in motion-jerkiness or ghost artifacts. As such, interpolators that consider the motion between frames are preferred, often times leading to pleasant conversion. The widely used block/patch-based methods consist of block-matching step that seek for similar blocks in a limited search window within the consecutive (and known) frames, leading to an estimation of the local motion trajectory. Then, new interpolated frames are obtained by compensating the motions of the matched blocks.

Similarly to previous applications, the effectiveness of the context is tested by integrating the context-feature only to the block-matching step. Therefore, a gain in performance is achieved due to improved matching, leading to better local motion estimation than the conventional block-based approach. In this experiment, we build upon the popular bi-directional technique \cite{choi2000new,zhai2005low,vinh2009frame}, which estimates the motion of $ 16\times 16 $ block in half-pixel accuracy, within a search radius of $ 10 $ pixels. The context-feature is computed by measuring the similarity between $ 7\times 7 $ patch to its $ 21 \times 21 $ neighborhood ($ \sigma = 10 $, $ m=4 $), concatenated to the conventional $ 16 \times 16 $ block, where $ \sqrt{\alpha} = 1.3 $.

Table \ref{fruc_compare} demonstrates the effect of the context on the bi-directional MC-FRUC for various video sequences, which are commonly used in related publications. Per each sequence, we measure the average PSNR over 80 interpolated frames, temporally up-sampled by a factor of 2. Following the results in Table \ref{fruc_compare}, the average improvement (over all the sequences) that achieved by plugging the context-feature to the conventional-blocks is 0.11dB, expressing the ability of the context to improve the block-matching step. 

A visual illustration can be found in Fig. \ref{fruc_results}, presenting the estimation of a frame from \textsf{Foreman} sequence. More specifically, given the existing frames at times $ t-1 $ (\mbox{Fig. \ref{fruc_results}a}) and $ t+1 $ (Fig. \ref{fruc_results}c), our goal is to estimate the unknown frame at time $ t $ (Fig. \ref{fruc_results}b). As can be seen, simple averaging of frames $ t-1 $ and $ t+1 $ leads to ghost artifacts (\mbox{Fig. \ref{fruc_results}d}), while the conventional motion-compensated approach results in much better performance (Fig. \ref{fruc_results}e). Still, due to errors in the block-matching step, it suffers from visible artifacts/outliers in the reconstruction (e.g. misaligned block close to the mouth). The context-based approach (Fig. \ref{fruc_results}f) benefits from better block-matching, leading to better estimation of the motion trajectory, and thereby resulting in improved restoration. The achieved improvement is also demonstrated in Fig. \ref{fruc_results}g and Fig. \ref{fruc_results}h that show the absolute difference between the interpolated and the original frame, for the conventional and context-based approach (linearly scaled to 0-255 range).

To summarize this experiment, we evaluate the effectiveness of working with context-blocks/patches as a way to improve temporal interpolation. Our experiments show that an improved performance can be achieved by integrating the context-feature to the block-matching step in MC-FRUC, supporting the effectiveness of the context.

\section{Conclusions}
\label{conclusions}

The motivation of this paper emerged from the impressive work of Levin et al. \cite{levin2011natural,levin2012patch} that suggest a way to evaluate the denoising bound of image patches. Theoretically, their work claims that larger patches result in better denoising performance compared to using small ones. Nevertheless, in order to ensure meaningful matches, the size of the database (number of samples) should grow exponentially with the patch-size quickly leading to unrealistic sizes. A "dense-sampling" assumption of the patch space holds true for small patches, while becoming unattainable when working with large patches.

These observations led us to present the con-patch: A low dimensional vector that benefits from the information that large-patches provide, along with preserving the attractive properties of working with small-patches. This is done by concatenating a small patch to its context-feature -- a compact representation of the surroundings/context of the central patch. The proposed feature is obtained by measuring the similarity of a small patch to its neighborhood patches, organized as a normalized histogram. As such, it represents the empirical distribution of the co-occurrences of the small central patch in its large surrounding window.

We tested the effect of the con-patches on several very distinct image processing applications that involve a NN search as their core mechanism. More specifically, the robustness of the con-patches to noise was demonstrated via image denoising. The ability to find meaningful NNs despite missing of spatial information is shown by treating the image SR problem, and better estimation of motion-trajectories is evaluated by integrating the context in MC-FRUC.
 
We note that our choice of features, while shown to be effective, is rather arbitrary. Further work is required to better design those features, perhaps by learning them.

\section*{Acknowledgment}
The authors would like to thank Jeremias Sulam and Vardan Papyan for their fruitful discussions. This research was supported by the European Research Council under EU's 7th Framework Program, ERC Grant agreement no. 320649.




\bibliographystyle{IEEEtran}
\bibliography{egbib}

\end{document}